\pdfoutput=1
\documentclass[acmsmall,nonacm]{acmart}
\usepackage{algorithmic}
\usepackage{algorithm}
\usepackage{multirow}
\usepackage{booktabs}
\usepackage{makecell}
\usepackage{color}
\AtBeginDocument{%
  }

\setcopyright{none}

\begin{document}

\title{Task-Relevant Representation Decoupling for Visual Reinforcement Learning Generalization}

\author{Jinwen Wang}
\email{jw.wang@bjtu.edu.cn}
\orcid{0009-0004-6280-0801}
\author{Youfang Lin}
\email{yflin@bjtu.edu.cn}
\orcid{0000-0002-5143-3645}
\author{Xiaobo Hu}
\email{xiaobohu@bjtu.edu.cn}
\orcid{0000-0001-6541-2784}
\author{Qian Xu}
\email{22125241@bjtu.edu.cn}
\orcid{0009-0002-6574-6092}
\author{Shuo Wang}
\email{shuo.wang@bjtu.edu.cn}
\orcid{0000-0001-6599-3638}
\affiliation{%
  \institution{Beijing Key Laboratory of Traffic Data Mining and Embodied Intelligence, School of Computer Science \& Technology, Beijing Jiaotong University}
  \country{China}
}

\author{Zhuo Chen}
\email{huntercz@126.com}
\orcid{0009-0001-4222-6601}
\affiliation{%
  \institution{CSSC Intelligent Innovation Research Institute}
  \country{China}
}
\affiliation{%
  \institution{Zhejiang University}
  \country{China}
}

\author{Kai Lv}
\email{lvkai@bjtu.edu.cn}
\orcid{0000-0001-6533-5176}
\authornote{Corresponding author: Kai Lv
}
\affiliation{%
  \institution{Beijing Key Laboratory of Traffic Data Mining and Embodied Intelligence, School of Computer Science \& Technology, Beijing Jiaotong University}
  \country{China}
}

\renewcommand{\shortauthors}{Jinwen Wang et al.}

\begin{abstract}
Visual Reinforcement Learning (VRL) has achieved considerable success in solving control tasks.
However, generalizing learned policies to new environments remains a major challenge, as agents often overfit to task-irrelevant features in the training environment. 
To solve this problem, we introduce the concept of decoupling observations into task-relevant and task-irrelevant representations.
Building on this idea, we propose a self-supervised \textbf{T}ask-\textbf{R}elevant \textbf{R}epresentation \textbf{D}ecoupling (T2RD) algorithm for VRL. 
This algorithm consists of three components: \textit{task-relevant representation consistency}, \textit{cross-reconstruction}, and \textit{cross-dynamic prediction}.
The first two components achieve the decoupling of content and style features, but the resulting content representations are not necessarily task-relevant.
To further refine task-relevant features from content representations, we design the third component that introduces dynamic prediction.
T2RD achieves State-Of-The-Art (SOTA) generalization performance and sample efficiency in the DeepMind Control Suite and Robotic Manipulation tasks.
\end{abstract}




\maketitle

\section{Introduction}
Visual Reinforcement Learning (VRL) has achieved tremendous success in various fields, including video games~\cite{DBLP:journals/corr/abs-2306-12860,DBLP:journals/nature/MnihKSRVBGRFOPB15}, robotic manipulation~\cite{DBLP:journals/corr/abs-2309-10150,DBLP:journals/corr/abs-2307-05973,DBLP:conf/mm/WangLHYH0025}, and autonomous navigation~\cite{DBLP:journals/jmlr/LevineFDA16, 10041739}.
However, high-dimensional image observations inherently contain redundant, task-irrelevant details, which makes the agent prone to overfitting the training environment. 
This poses significant challenges in generalizing learned policies to new environments with visual changes. 
Therefore, it is crucial to improve the generalization ability of VRL. 

To learn policies robust to visual changes, previous methods can be broadly categorized into two types.
One is data augmentation ~\cite{DBLP:conf/nips/LaskinLSPAS20, DBLP:journals/corr/abs-2004-13649, DBLP:conf/icml/FanWHY0ZA21, DBLP:conf/ijcai/YuanMMX00LX22, DBLP:journals/corr/abs-2306-00656}, aiming to cover sufficient environmental changes during training that also exist in the testing environment.
The other is designing self-supervised auxiliary tasks~\cite{DBLP:conf/nips/GrillSATRBDPGAP20, DBLP:journals/corr/abs-2011-13389, DBLP:journals/corr/abs-2312-01915, DBLP:conf/icml/LaskinSA20, DBLP:conf/iclr/0001MCGL21, hu2025bidirectional}  to assign certain properties to the representation.
However, representations obtained from previous self-supervised auxiliary tasks don't explicitly strip away task-irrelevant information.
The policies based on these representations may also overfit to task-irrelevant features, thereby affecting policies' generalization ability. 

\begin{figure}[t]
  \centering
  \includegraphics[width=0.65\linewidth]{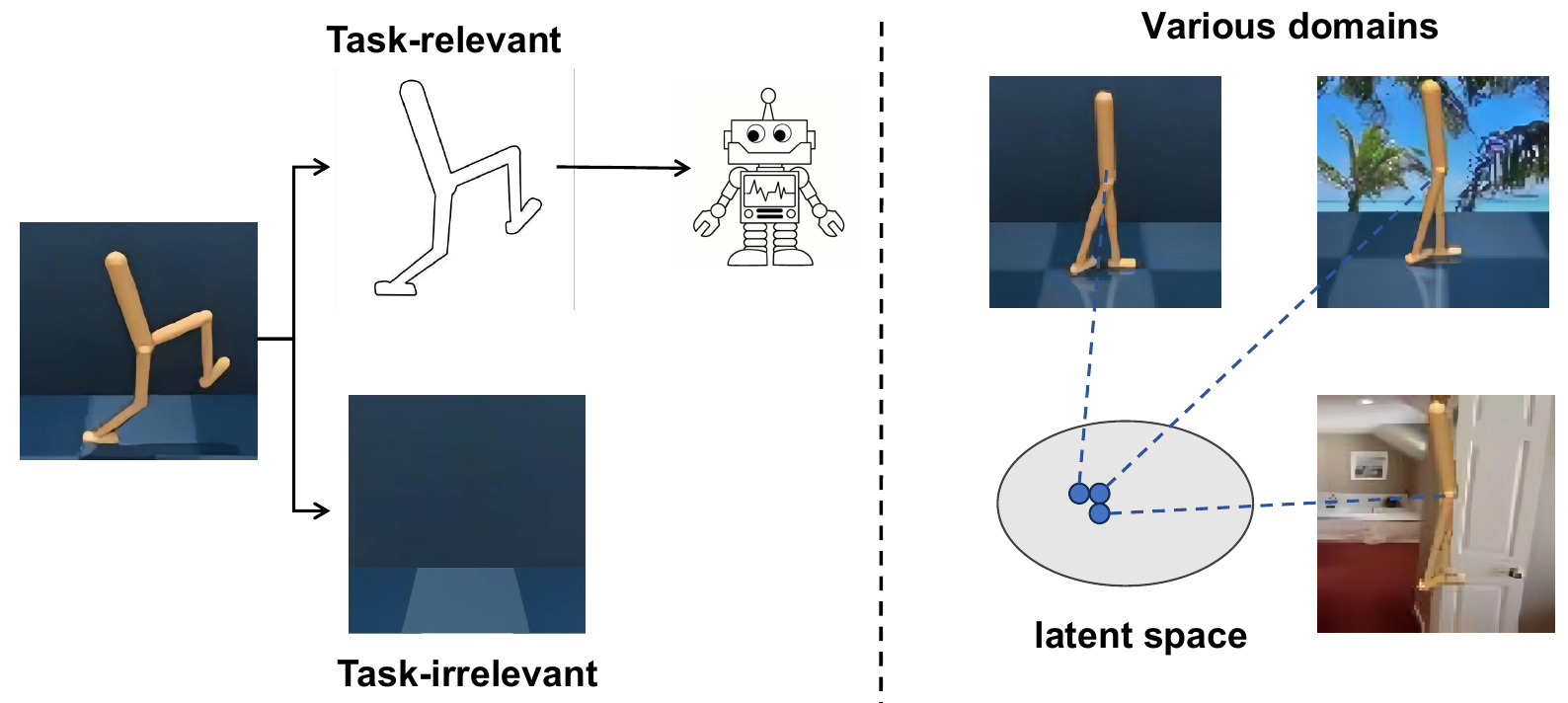}
  \caption{\textbf{The core idea of our method.} We decouple the visual observation into task-relevant representation and task-irrelevant representation. Task-relevant representations have domain-invariant properties, allowing them to generalize well across different domains.}
  \Description{}
\label{fig:intro}
\end{figure}

To address this problem, as shown in Figure ~\ref{fig:intro}, we propose the idea of decoupling the observations into task-relevant and task-irrelevant representations. 
Then, the task-relevant representations are utilized to support learning policies. 
This decoupling idea has two advantages: 1) The representations supporting policy learning are expected to exclude task-irrelevant information, making policies focus more on the task-relevant parts of the observations and thus promoting policy generalization;
2) By explicitly ignoring task-irrelevant features, the burden on the agent to process unnecessary information is reduced, thereby improving sample efficiency.

Based on the idea mentioned above, we introduce the Task-Relevant Representation Decoupling (T2RD) method for VRL, which is capable of zero-shot generalization to unseen visual scenarios. 
Specifically, to achieve the decoupling effect, T2RD consists of three components: \textit{task-relevant representation consistency}, \textit{cross-reconstruction}, and \textit{cross-dynamic prediction}. 
The first two components are designed to separate content and style features visually. However, the resulting content representations may not necessarily be task-relevant. 
To further refine task-relevant features from the content representation, we designed the third component that integrates dynamic prediction, thereby achieving the decoupling of task-relevant and task-irrelevant features.

T2RD has two principal advancements: 1) Its representation decoupling module is self-supervised.
Although there are similar decoupling ideas~\cite{DBLP:conf/nips/Gonzalez-Garcia18, DBLP:journals/neco/TenenbaumF00, DBLP:conf/nips/MathieuZZRSL16, DBLP:conf/eccv/SanchezSO20} in the field of Computer Vision (CV), most of them are supervised and require labeled datasets or even paired datasets.
2) It is more suitable for VRL tasks. 
Compared to methods that merely perform foreground-background segmentation \cite{DBLP:conf/cvpr/WangLY21} or content-style separation~\cite{DBLP:conf/iclr/YinRLWXZ22}, our decoupling approach is more applicable to generalization tasks in VRL, achieving separation between task-relevant and task-irrelevant features.

Additionally, we propose a simple yet effective combined data augmentation technique for VRL. 
Recent methods mostly select a single data augmentation technique~\cite{DBLP:journals/corr/abs-2011-13389, DBLP:conf/nips/HansenSW21,
DBLP:conf/ijcai/YuanMMX00LX22,
DBLP:journals/corr/abs-2312-01915}.
However, these methods suffer from generalization bias, meaning that their generalization ability is heavily influenced by what data augmentation techniques they choose.
By simply combining Random Convolution \cite{DBLP:journals/corr/abs-2011-13389} and Random Overlay \cite{DBLP:conf/nips/HansenSW21}, we increase the diversity of augmentations, alleviating the generalization bias and thereby improving the algorithm's generalization performance.

We conduct extensive experiments on two benchmarks: DeepMind Control Suite Generalization Benchmark (DMControl-GB) ~\cite{DBLP:conf/nips/HansenSW21, DBLP:journals/corr/abs-1801-00690} and Robotic Manipulation ~\cite{DBLP:journals/ral/JangirHGJW22}.
The agent is trained in a fixed environment and evaluated for its generalization ability to unseen environments during training. 
The experimental results demonstrate that our method exhibits state-of-the-art generalization performance and sample efficiency.

Our main contributions are emphasized as follows:
\begin{itemize}
    \item We introduce the concept of decoupling observations into task-relevant features and task-irrelevant features in VRL, and then use only task-relevant features for policy learning.
    \item We propose a self-supervised representation decoupling framework suitable for VRL, which can effectively achieve the decoupling of task-relevant features and task-irrelevant features.
    \item We propose a simple yet effective combined data augmentation technique, which significantly improves the generalization performance of VRL, as evidenced by experiments.
    \item Extensive experiments prove that T2RD achieves state-of-the-art generalization performance and sample efficiency.
\end{itemize}

\section{RELATED WORK}
\subsection{Data Augmentation in Visual RL}
Data augmentation has been shown to improve the generalization ability of visual reinforcement learning algorithms. 
RAD ~\cite{DBLP:conf/nips/LaskinLSPAS20} pioneers the broad application of data augmentation in RL, improving sample efficiency and generalization. 
DrQ ~\cite{DBLP:journals/corr/abs-2004-13649} aggregates multiple data augmentations, regularizing learning directly from pixels by averaging across multiple image transformations for the Q-function and Q target function. 
SODA ~\cite{DBLP:journals/corr/abs-2011-13389} proposes a soft data augmentation method that decouples data augmentation from policy learning and introduces a novel data augmentation approach that linearly adds observations with additional images. 
TLDA ~\cite{DBLP:conf/ijcai/YuanMMX00LX22} introduces task-aware Lipschitz data augmentation, explicitly identifying task-relevant pixels with large Lipschitz constants and augmenting only task-irrelevant pixels. 
Compared with single data augmentations, our combined data augmentation method increases the diversity of augmentation, which can alleviate the issue of generalization bias. 
At the same time, using different augmentations improves the network's robustness to visual changes.

\subsection{Generalization in Visual RL}
Researchers have studied how to improve the generalization ability of visual reinforcement learning from various aspects. 
Beyond data augmentation, existing works primarily pursue generalization by learning invariant representations, which can be broadly categorized into three main approaches.
First, one category uses self-supervised objectives to learn representations. 
These methods impose auxiliary self-supervised tasks on visual data to encourage the model to learn general features beneficial for the task. 
Representative works include using contrastive learning (CURL~\cite{DBLP:conf/icml/LaskinSA20}), consistency regularization (SODA~\cite{DBLP:journals/corr/abs-2011-13389}), domain-invariant representation learning~\cite{DBLP:conf/ijcai/Wang0WHLL24}, or foreground/saliency extraction (VAI~\cite{DBLP:conf/cvpr/WangLY21}, SGQN~\cite{DBLP:conf/nips/BertoinZZR22}).
Second, another category leverages the intrinsic structure of the Markov Decision Process (MDP) and RL characteristics. 
This includes using bisimulation metrics~\cite{DBLP:journals/siamcomp/FernsPP11} to group behaviorally similar states (DBC~\cite{DBLP:conf/iclr/0001MCGL21}, Chen et al.~\cite{DBLP:conf/nips/ChenP22}), achieving policy adaptation after deployment via inverse dynamics prediction (PAD~\cite{DBLP:conf/iclr/HansenJSAAEPW21}), stabilizing Q-learning through asymmetric data augmentation (SVEA~\cite{DBLP:conf/nips/HansenSW21}),
bidirectionally predicting environment dynamics (BiT~\cite{DBLP:journals/corr/abs-2312-01915}), leveraging long-term information~\cite{DBLP:conf/iclr/YangWPGW025}, or designing temporal contrastive losses~\cite{DBLP:conf/nips/ZhengWSMZXDH23}. 
Other strategies involve robustness-oriented learning and analysis~\cite{DBLP:conf/icml/ZhongSLY021, DBLP:journals/pami/LuoSZLZW20, zhang2024certified} and designing intrinsic reward systems to achieve self-supervised domain adaptation~\cite{DBLP:conf/cvpr/ZuoQXZWY19}.
While these methods effectively learn a single, robust representation by implicitly ignoring distractors, T2RD's philosophy is different. 
We argue for explicitly decoupling an observation into task-relevant and task-irrelevant streams, then using dynamic prediction to ensure the task-relevant stream is behaviorally meaningful, which is a more direct path.
Third, a powerful paradigm utilizes pre-trained Visual Foundation Models (VFMs). 
For instance, some works use SAM for object segmentation to remove background distractors~\cite{DBLP:journals/corr/abs-2312-17116}, or directly employ general VFM features as input (PIE-G~\cite{DBLP:conf/nips/YuanXYWWGX22}, Zhong et al.~\cite{DBLP:conf/eccv/ZhongWCWC24}). 
However, the features of the segmented objects themselves—their color, texture, or lighting—can still be task-irrelevant and cause overfitting. 
T2RD is positioned to address this problem by further decoupling an object's task-relevant state from its superficial appearance.

\subsection{Representation Decoupling}
Human perceptual systems often distinguish between content and style to better understand observations~\cite{DBLP:journals/neco/TenenbaumF00}, and in artificial intelligence, decomposed representations of content and style are also highly needed. 
In computer vision, for example, Lorenz et al.~\cite{DBLP:conf/cvpr/LorenzBMO19} separate appearance and shape by learning consistent parts across instances of the same category, while Wu et al.~\cite{DBLP:conf/iclr/WuCLQL19} introduce a two-branch autoencoder to disentangle objects into content and style representations. 
Ren et al.~\cite{DBLP:conf/iccvw/RenYWZ21} propose a content-style disentanglement module by encoding dominant image factors. 
Yin et al.~\cite{DBLP:conf/iclr/YinRLWXZ22} model the content-style representation as a token-level bipartite graph.

This principle of learning informative representations also extends to visual recognition and spatio-temporal modeling~\cite{fan2018unsupervised,fan2022point}, as well as medical image analysis~\cite{DBLP:conf/icml/WangGZZ0W22}, and a related line of work leverages direct human perceptual cues to guide agents;
for example, human gaze data can direct an agent's attention for efficient imitation learning~\cite{DBLP:conf/atal/SaranZSN21}, and human-generated saliency has been used to guide value estimation in visual RL~\cite{DBLP:conf/iros/LiangTB24}. While these methods validate the benefit of isolating task-relevant information, they share a common limitation with most computer vision decoupling approaches: reliance on external supervision, whether paired datasets, labeled medical data, or direct human input. T2RD is distinct in that it learns this decoupling in a self-supervised manner tailored for VRL, using only the standard reward signal without external labels or human guidance.

\section{Preliminaries}

\subsection{Visual Reinforcement Learning}

Due to the partial observability of images in Visual Reinforcement Learning (VRL), the interaction between an agent and the environment is modeled as a Partially Observable Markov Decision Process (POMDP), represented by the tuple
$\mathcal{M} = \langle \mathcal{S}, \mathcal{O}, \mathcal{A}, \mathcal{P}, \mathcal{Z}, r, \mathcal{\gamma} \rangle$,
where 
$\mathcal{S}$ is the state space, 
$\mathcal{O}$ is the high-dimensional observation space, 
$\mathcal{A}$ is the action space, 
$\mathcal{P}(\mathbf{s}_{t+1} \mid \mathbf{s}_{t}, a_{t})$ 
is the state transition function,
$\mathcal{Z}(\mathbf{o}_{t} \mid \mathbf{s}_{t})$ is the observation-emission distribution,
$r:\mathcal{S}\times \mathcal{A}\mapsto  \mathbb{R} $ is the scalar reward function, 
and $\gamma \in \left [  0,1 \right ) $ is the discount factor.

The objective is to learn the optimal policy $\pi^*$ that maximizes the cumulative reward return:
\begin{equation}
\pi^* = \arg \max_{\pi} \;
\mathbb{E}_{\,\mathbf{o}_t \sim \mathcal{Z}(\cdot \mid \mathbf{s}_t),\; a_t \sim \pi(\cdot \mid \mathbf{o}_t),\; \mathbf{s}_{t+1} \sim \mathcal{P}(\cdot \mid \mathbf{s}_t, a_t)}
\Big[ \sum_{t=1}^{T} \gamma^t \, r(\mathbf{s}_{t}, a_{t}) \Big]
\end{equation}
starting from the initial state $\mathbf{s}_0 \in \mathcal{S}$ and taking actions 
$a_t$ chosen by the observation-conditional policy $\pi_{\theta}(\cdot \mid \mathbf{o}_t)$, 
while observations are generated via $\mathcal{Z}(\mathbf{o}_t \mid \mathbf{s}_t)$ and the environment transitions to $\mathbf{s}_{t+1}$ according to $\mathcal{P}(\cdot \mid \mathbf{s}_t, a_t)$.
The policy is parameterized by learnable parameters $\theta$.




\subsection{Generalization in VRL}
In terms of generalization of visual reinforcement learning, we consider a set of POMDPs denoted by $\mathbf{M} = \{ \mathcal{M}_1, \mathcal{M}_2, \ldots, \mathcal{M}_n \}$, where each POMDP $\mathcal{M}_i$ has its own observation space $O_i$ but shares a common underlying state space $\mathcal{S}$ and state transition function $\mathcal{P}$.
During training, we only have access to a specific POMDP, denoted as $\mathcal{M}_i$. 
Our goal is to train an agent on a specific scenario $\mathcal{M}_i$ to learn a policy $\pi^*_G$ that maximizes the expected cumulative reward across the entire set of POMDPs $\mathbf{M}$, in a zero-shot generalization manner.

\begin{figure}
  \centering
  \includegraphics[width=0.88\linewidth]{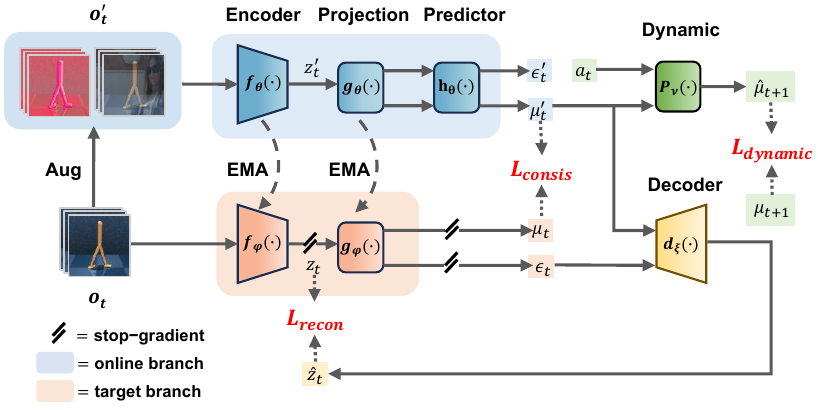}
  \caption{\textbf{Overview of our model.} Our model includes two branches: the online branch and the target branch. 
T2RD consists of three components: \textit{task-relevant representation consistency}, \textit{cross-reconstruction}, and \textit{cross-dynamic prediction}. 
Our objective is to achieve the decoupling of task-relevant and task-irrelevant features through the above three modules.
Subsequently, only the task-relevant features are used for policy learning.
}
\label{fig:Overview}
\end{figure}

\section{METHOD}

\subsection{Overview}
As shown in Figure~\ref{fig:Overview}, we propose Task-Relevant Representation Decoupling (T2RD). 
Our architecture is inspired by successful self-supervised learning methods such as BYOL~\cite{DBLP:conf/nips/GrillSATRBDPGAP20} and adopts a dual-branch design: an online branch and a target branch. The core idea is that the online branch is trained to learn from the representations produced by the target branch, which evolves slowly and thus provides a stable regression target. This setup mitigates training instability and helps prevent collapse to trivial solutions.

The online branch comprises three components, each serving a distinct purpose: an encoder $f_{\theta}$ that extracts latent representations from pixels, a dual-head projection $g_{\theta}$ that decouples the representation into task-relevant content and task-irrelevant style, and a predictor head $h_{\theta}$ that learns to predict the target-branch representation from the online-branch representation. 
By requiring a non-trivial mapping between the online and target branches, this predictor helps avoid representation collapse.
The target branch has a structurally identical encoder $f_{\varphi}$ and dual-head projection $g_{\varphi}$, but no predictor.

For policy learning, we adopt Soft Actor-Critic (SAC)~\cite{DBLP:conf/icml/HaarnojaZAL18}. As shown in Figure~\ref{fig:Policy Learner}, the T2RD module serves as a representation learner: we first decouple visual inputs into task-relevant and task-irrelevant features, then feed only the task-relevant representations to SAC.



\subsection{Self-Supervised Decoupling of Latent Representations}
Our decoupling process begins with data preparation and augmentation. During training, we sample a batch of transitions $\{o_t, a_t, r_t, o_{t+1}\} \sim \mathcal{B}$ from a replay buffer $\mathcal{B}$. For each observation $o_t$, we generate an augmented view $o_t' = t(o_t)$ using our combined data augmentation, which samples a transformation $t$ with equal probability from a set of augmentations $\mathcal{T}$ (containing Random Overlay and Random Convolution). As depicted in Figure~\ref{fig:Overview}, the augmented observation $o_t'$ is subsequently processed by the online branch, while the original observation $o_t$ is fed into the target branch.

A crucial element of this dual-branch setup is preventing the model from collapsing. This is achieved by managing the target network's updates. We apply a stop-gradient to the target branch's output, preventing gradients from flowing into it. Instead of being trained via backpropagation, its parameters $\varphi$ are updated as a slow-moving average of the online network's parameters $\theta$. This Exponential Moving Average (EMA) update ensures the target network provides a stable and consistent objective, which is essential for stabilizing the entire self-supervised training process:
\begin{equation}
  \varphi_{n+1} \leftarrow (1 - \rho) \varphi_n + \rho \theta_n,
\end{equation}
where $\rho$ is the momentum coefficient for the EMA update, with its specific value listed in the hyperparameters of Table~\ref{table:hyperparameters}.
Our self-supervised representation learning process is then driven by three key objectives, which we detail below.

\subsubsection{Task-Relevant Representation Consistency.}
The first objective ensures that the core ``content'' of an observation remains consistent even when its ``style'' is changed by data augmentation. The augmented observation \(o_{t}'\) and the original observation \(o_{t}\) share the same underlying state. Therefore, their task-relevant representations should be close. 
Thus, the content features \({\mu}_{{t}}\), \({\mu}_{{t}}'\)  extracted from \({o}_{{t}}\) and \({o}_{{t}}'\) should be consistent. 
The objective function of \textit{task-relevant representation consistency} can be expressed as:
\begin{equation}
L_{\text{consis}} = \mathbb{E}_{\tau \sim D_{\text{env}}} \left[\| \mu_{t}' - \mu_{t} \|_2^2 \right],
\label{Equ2}
\end{equation}
where $D_{\text{env}}$ is the replay buffer. 
Equation~\ref{Equ2} ensures the consistency of content features between the original and augmented images, thereby excluding style-related information from the obtained content representations $\mu$. Such content representations possess domain-invariant properties, facilitating better generalization across different domains.

\subsubsection{Cross-Reconstruction for Decoupling.}
To decouple content and style, we introduce a reconstruction objective. The logic is as follows: while the consistency loss in Equation~\ref{Equ2} drives the content representation $\mu$ to become style-invariant, we must also ensure information sufficiency and separability.
The reconstruction objective guarantees that the combination of $\mu$ and $\epsilon$ preserves the complete information of the original observation $o$. 
Thus, as $\mu$ is purified of style, the style information is naturally compressed into $\epsilon$.

Building on this, we introduce a novel twist to the standard reconstruction approach: \textbf{cross-reconstruction}. 
This technique forces a cleaner and more robust separation of content and style.
Specifically, we reconstruct the full latent representation $z_t$ of the original observation by combining the content representation $\mu_t'$ from the augmented view with the style representation  $\epsilon_t$ from the original view:
\begin{equation}
\hat{z}_t = d_\xi(\mu_{t}', \epsilon_t),
\label{Equ3}
\end{equation}
The \textit{cross-reconstruction} objective then minimizes the difference between the reconstructed $\hat{z}_t$ and the original $z_t$:
\begin{equation}
L_{\text{recon}} = \mathbb{E}_{{\tau} \sim D_{\text{env}}} \left[\| z_t - \hat{z}_t \|_2^2 \right].
\label{Equ4}
\end{equation}
The ``cross'' design is critical: if the content representation $\mu_t'$ still contains any style information from the augmented view, it will conflict with the style $\epsilon_t$ from the original view, leading to a poor reconstruction and a high loss. This incentivizes the model to learn a pure, decoupled content representation.

\subsubsection{Dynamic Prediction for Task-Relevance.}
The first two components successfully decouple observations into a style-invariant ``content'' stream and a ``style'' stream. 
However, the obtained content representation is not guaranteed to be task-relevant. 
For example, the agent's shadow might be considered content, but it is irrelevant for controlling the agent. 
To solve this, we introduce a dynamic prediction module. The intuition is that a truly task-relevant representation must contain information that is predictive of future states under the agent's actions. 
By enforcing this predictive constraint, we refine the content representation to be specifically task-relevant. 
This crucial step elevates our framework from a simple content-style separator to a true decoupler of task-relevant and task-irrelevant features.

We also implement a \textbf{cross-domain prediction} technique here. Specifically, we use the task-relevant representation ($\mu_t'$) from the augmented view and the agent's action ($a_t$) to predict the task-relevant representation ($\mu_{t+1}$) of the \textit{original} next state:
\begin{equation}
\hat{\mu}_{t+1} = P_v(\mu_{t}', a_{t}).
\label{Equ5}
\end{equation}
This cross-domain objective is a powerful regularizer for generalization. It forces the representation $\mu$ to be not only predictive but also domain-invariant, as it must bridge the visual gap between the augmented and original observation domains to succeed. 
The \textit{cross-dynamic prediction} objective is:
\begin{equation}
L_{\text{dynamic}} = \mathbb{E}_{{\tau} \sim D_{\text{env}}} \left[\| \hat{\mu}_{t+1} - \mu_{t+1} \|_2^2 \right].
\label{Equ6}
\end{equation}

\subsubsection{Overall Objective.}
We jointly optimize the three objectives. The overall objective function for our self-supervised module is:
\begin{equation}
L_{T2RD} = \lambda_1 \cdot L_{\text{consis}} + \lambda_2 \cdot L_{\text{recon}} + \lambda_3 \cdot L_{\text{dynamic}}.
\end{equation}
Here, $\lambda_1$, $\lambda_2$, and $\lambda_3$ are scalar coefficients (with $\lambda_i>0$) that balance the contribution of each loss term. The specific values used in our experiments are summarized in Table~\ref{table:hyperparameters}.

\begin{figure}
  \centering
  \includegraphics[width=0.6\linewidth]{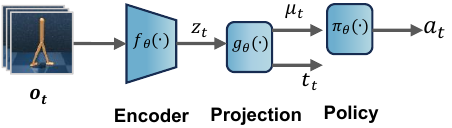}
  \caption{\textbf{Policy Learner.} The observation \({o}_{{t}}\) is decoupled into the task-relevant representation \( \mu_t \) and the task-irrelevant representation \( \epsilon_t \) through an encoder \( {f}_{{{\theta}}} \) and a dual-head projection \( {g}_{{{\theta}}} \). 
  Then, only the task-relevant representation \( \mu_t \) is utilized for policy learning.}
\label{fig:Policy Learner}
\end{figure}

\begin{algorithm}[tb]
\caption{\textbf{T2RD}}
\begin{algorithmic}[1]
\STATE ${\theta},{\varphi},{v},{\xi} $: randomly initialized decoupling parameters\;
\STATE ${\omega}$: RL updates per iteration\;
\STATE $\rho$: momentum coefficient\;
\FOR{every iteration}
    \FOR{update = 1, 2, ..., $\omega$}
        \STATE Sample batch of transitions $\{o_t, a_t, r_t, o_{t+1}\} \sim \mathcal{B}$
        \STATE Obtain $\mu_t$ by ${f}_{{{\theta}}}$ and ${g}_{{{\theta}}}$
        \STATE Minimize $L_{RL}$
    \ENDFOR
    \STATE Sample batch of transitions $\{o_t, a_t, r_t, o_{t+1}\} \sim \mathcal{B}$
    \STATE Augment observations $o_t' = Aug(o_t)$
    \STATE Obtain ${\mu}_t', {\epsilon}_t' = {h}_{{\theta}}({g}_{{{\theta}}}({f}_{{{\theta}}}({o}_t')))$
    \STATE Obtain ${\mu}_t, {\epsilon}_t = {g}_{{\varphi}}({f}_{{\varphi}}({o}_t))$
    \STATE Obtain $\mathbf{\mu}_{t+1}, \epsilon_{t+1} = {g}_{{\varphi}}({f}_{{\varphi}}({o}_{t+1}))$
    \STATE Cross reconstruction $\hat{z}_t = d_\xi({\mu}_t', {\epsilon}_t)$
    \STATE Dynamic prediction $\hat{\mathbf{\mu}}_{t+1} = P_v({\mu}_t', {a}_t)$
    \STATE Consistency loss $L_{\text{consis}} = \mathbb{E}_{\tau \sim D_{\text{env}}} \left[\|{\mu}_t' - {\mu}_t \|_2^2 \right]$
    \STATE Reconstruction loss $L_{\text{recon}} = \mathbb{E}_{\tau \sim D_{\text{env}}} \left[\| {z}_t - \hat{z}_t \|_2^2 \right]$
    \STATE Dynamic loss $L_{\text{dynamic}} = \mathbb{E}_{\tau \sim D_{\text{env}}} \left[\| \hat{\mu}_{t+1} - {\mu}_{t+1} \|_2^2 \right]$
    \STATE Optimize $L_{T2RD}$ w.r.t. $\theta, v, \xi$
    \STATE Update $\varphi \leftarrow (1 - \rho) \varphi + \rho \theta$
\ENDFOR
\end{algorithmic}
\label{alg:T2RD}
\end{algorithm}

\subsection{Policy Learner}
T2RD, as a representation decoupling module, is independent of the policy learner and applicable to any visual reinforcement learning (RL) algorithm.
We use the Soft Actor-Critic (SAC) algorithm for policy learning.
SAC is an actor-critic RL approach based on the maximum entropy framework, concurrently learning the Q function and the policy network \( \pi_{{\theta}} \), to maximize the expected cumulative reward.
Specifically, as shown in Figure ~\ref{fig:Policy Learner}, the observation \({o}_{{t}}\) is decoupled into the task-relevant representation \(\mu_t\) and task-irrelevant representation \(\epsilon_t\) through the encoder \({f}_{{\theta}}\) and the dual-head mapping \({g}_{{\theta}}\).
Then, the policy learner uses only the task-relevant representation \(\mu_t\) for policy learning. 
As shown in Algorithm~\ref{alg:T2RD}, the optimization of representation learning \( L_{T2RD} \) alternates with the optimization of reinforcement learning \( L_{RL} \) with \( {\omega} \) steps. 
The two optimization objectives jointly update the parameter \( \theta \). 

\subsection{Discussions}
1) \textit{What sets T2RD apart from other self-supervised auxiliary methods?} We explicitly aim to decouple task-relevant and task-irrelevant features in the observation, where the encoding processes of task-relevant and task-irrelevant features complement each other. 
This deepens the understanding of visual observations, thereby obtaining more precise task-relevant representations. 
For instance, knowing what is wrong can lead to a clearer understanding of what is right.
Furthermore, during policy learning, by explicitly discarding task-irrelevant features, we can reduce the burden of information processing on the agent, thereby improving sample efficiency.

2) \textit{Why does T2RD greatly benefit from our combined data augmentation?}
As shown in the subsequent experimental section, while other algorithms also benefit from our combined data augmentation, the improvements are not as substantial as those seen with T2RD. 
We argue that while the combined data augmentation increases augmentation diversity, it may also lead to some interference and conflicts among them.
Thanks to T2RD's robust decoupling capabilities, it can fully exploit this increased diversity, maximizing the advantages of the combined data augmentation.

\subsection{Theoretical Analysis}

To ground our framework theoretically, we connect our approach to \textbf{bisimulation metrics}~\cite{DBLP:journals/siamcomp/FernsPP11, DBLP:conf/iclr/0001MCGL21}, which quantify behavioral similarity between states in an MDP. 
The metric $d_{\mathcal{M}}(s_i, s_j)$ is defined as the fixed point of the following Bellman-like operator:
\begin{equation}
\label{eq:bisim}
d_{\mathcal{M}}(s_i, s_j) = \max_{a \in \mathcal{A}} \left( |r(s_i, a) - r(s_j, a)| + \gamma W_1(\mathcal{P}(\cdot|s_i, a), \mathcal{P}(\cdot|s_j, a); d_{\mathcal{M}}) \right).
\end{equation}
where $W_1$ is the 1-Wasserstein distance measuring the difference between next-state transition distributions. 
Crucially, this metric provides an upper bound on the optimal value function difference:

\textbf{Theorem 1} (Ferns et al.~\cite{DBLP:journals/siamcomp/FernsPP11}). For any two states $s_i, s_j \in \mathcal{S}$, the difference in their optimal value functions is bounded by their bisimulation distance:
\begin{equation}
\label{eq:ferns_bound}
|V^*(s_i) - V^*(s_j)| \le d_{\mathcal{M}}(s_i, s_j).
\end{equation}
This provides a clear analytical path: if T2RD learns a representation $\mu$ where the latent distance $||\mu(o_i) - \mu(o_j)||_2$ approximates the bisimulation distance $d_{\mathcal{M}}$, we can then bound the error of a value function learned upon this representation.
Our complete objective is designed to provide a strong inductive bias for learning such a metric from high-dimensional observations.

\subsubsection{Connecting T2RD Losses to the Bisimulation Metric}
We first map high-dimensional observations to states and then establish the connection with the bisimulation metric.


\textbf{State Invariance and Sufficiency ($L_{\text{consis}}$ \& $L_{\text{recon}}$).} The bisimulation metric is defined over underlying states $s$, while our agent sees observations $o$. Therefore, the representation must be invariant to task-irrelevant visual features while remaining sufficient information for control.

(1) The \textbf{consistency loss ($L_{\text{consis}}$)} enforces \textbf{state invariance}. Since an observation $o_t$ and its augmentation $o_t'$ correspond to the same underlying state $s_t$, their task-relevant representations should be identical. 
Minimizing $||\mu_{t}^{\prime}-\mu_{t}||_{2}^{2}$ achieves precisely this.

(2) The \textbf{cross-reconstruction loss ($L_{\text{recon}}$)} ensures \textbf{information sufficiency and separation}. 
First, reconstruction guarantees that the learned representations ($\mu$, $\epsilon$) retain sufficient information from the high-dimensional observations $o$. Second, by forcing the decoder to reconstruct the original latent state $z_t$ from the \textit{augmented} task-relevant content $\mu_t'$ and the \textit{original} task-irrelevant style $\epsilon_t$, it ensures that the task-relevant content representation $\mu$ is purged of style information.
This mechanism acts as a practical information bottleneck, ensuring that the learned distance $||\mu_i - \mu_j||_2$ reflects differences between underlying states ($s_i, s_j$), not superficial visual distractors.

\textbf{Dynamics and Rewards ($L_{\text{dynamic}}$ \& $L_{\text{RL}}$).} The bisimulation metric in Equation~\ref{eq:bisim} has two components: reward difference and transition dynamics difference. Our framework addresses both.

(1) The \textbf{cross-dynamic prediction loss ($L_{\text{dynamic}}$)} encourages representations $\mu$ to respect the MDP's transition dynamics, acting as a practical, sample-based proxy for minimizing the Wasserstein distance term in the bisimulation metric.

(2) The \textbf{reward term} is implicitly modeled by the downstream policy learner ($L_{\text{RL}}$). 
As described in Algorithm~\ref{alg:T2RD}, the encoder's parameters $\theta$ are updated by gradients from both $L_{\text{T2RD}}$ and the reward-driven SAC loss $L_{\text{RL}}$~\cite{DBLP:conf/icml/HaarnojaZAL18}. 
The gradients from the SAC critic, which minimizes Bellman error, force the representation $\mu$ to distinguish between states that lead to different rewards.

This joint optimization ensures our learned representation $\mu$ is sensitive to both rewards and dynamics, thereby learning a functional approximation of the full bisimulation metric.

\subsubsection{Bound on the Action-Value Function}

Based on the argument that our learning objective guides the representation to satisfy the properties of a bisimulation metric, we can establish a bound on the action-value function's error.

\textbf{Theorem 2.} Under the assumption that the learned representation $\mu$ yields a distance metric $d_{\mu}(s_i, s_j) = ||\mu(o_i) - \mu(o_j)||_2$ that approximates the true bisimulation metric up to a bounded error $\epsilon_{\text{T2RD}}+ \epsilon_{\text{RL}}$ (i.e., $|d_{\mu}(s_i,s_j) - d_{\mathcal{M}}(s_i, s_j)| \le \epsilon_{\text{T2RD}}+ \epsilon_{\text{RL}}$), let $\hat{Q}(\mu(o), a)$ be the action-value function learned on the representations produced by an encoder trained with the combined T2RD and SAC objectives, and let $Q^*(s, a)$ be the true optimal action-value function for the underlying state $s$. The suboptimality is bounded as follows:
\begin{equation}
\label{eq:our_bound}
\max_{s, a} |Q^*(s, a) - \hat{Q}(\mu(o), a)| \le \frac{\epsilon_{\text{T2RD}} + \epsilon_{\text{RL}}}{1-\gamma}
\end{equation}
where $\epsilon_{\text{T2RD}}$ is the representation error, which our framework is designed to minimize via the $L_{\text{T2RD}}$ objective, and $\epsilon_{\text{RL}}$ is the Bellman error from the policy learner, controlled by the converged $L_{\text{RL}}$ loss.

\section{Experiments}
In this section, we conduct extensive experiments on \textit{DeepMind Control Suite Generalization Benchmark (DMControl-GB)} ~\cite{DBLP:journals/corr/abs-1801-00690} and \textit{Robotic Manipulation tasks} ~\cite{DBLP:journals/ral/JangirHGJW22} to validate the generalization performance and sample efficiency of our method.
The agent is trained in a fixed environment, and we evaluate the algorithm's generalization ability to unseen environments with more complex backgrounds or color variations.


\textbf{Baselines.}
We compare our approach with other SOTA algorithms:
1) \textbf{DrQ} ~\cite{DBLP:journals/corr/abs-2004-13649} applies data augmentation to both Q-values and target Q-values during TD updates.
2) \textbf{SODA} ~\cite{DBLP:journals/corr/abs-2011-13389} proposes soft data augmentation and employs a BYOL-like architecture to learn consistent representations.
3) \textbf{SVEA} ~\cite{DBLP:conf/nips/HansenSW21} only applies data augmentation to the Q values without augmenting target Q values.
4) \textbf{TLDA} ~\cite{DBLP:conf/ijcai/YuanMMX00LX22} introduces task-aware Lipschitz data augmentation.
5) \textbf{PIE-G} ~\cite{DBLP:conf/nips/YuanXYWWGX22} utilizes ImageNet ~\cite{DBLP:conf/cvpr/DengDSLL009} pre-trained encoders to provide a fairly general representation for RL.
6) \textbf{SGQN} ~\cite{DBLP:conf/nips/BertoinZZR22} learns masks through saliency maps and consistency regularization to exclude irrelevant pixels.
7) \textbf{BiT} ~\cite{DBLP:journals/corr/abs-2312-01915} introduces a bidirectional transition model.
8) \textbf{TRP} ~\cite{DBLP:conf/aaai/WangWHWLL24} introduces the truncated return prediction task, promoting cross-domain policy consistency.
9) \textbf{Visarl}~\cite{DBLP:conf/iros/LiangTB24} leverages human-generated saliency maps to guide visual RL, using human perceptual cues to focus learning on task-relevant regions.

\begin{figure}
  \centering
  \includegraphics[width=0.45\linewidth]{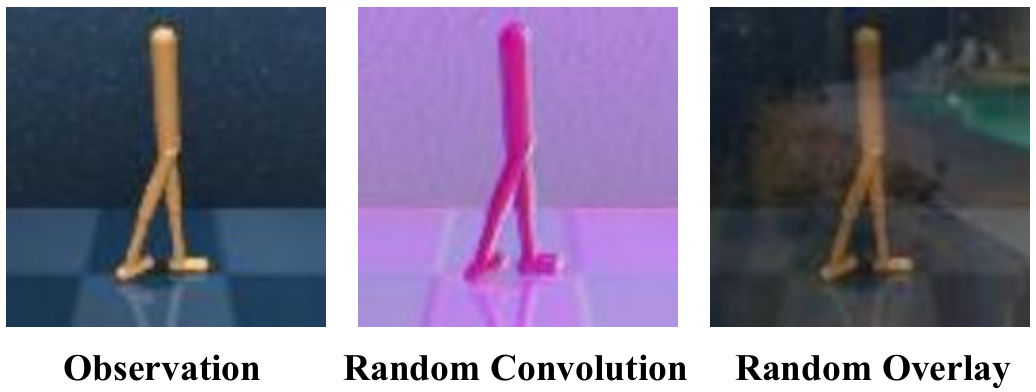}
  \caption{\textbf{Examples of data augmentation.}
  We separately present the original observation and the results after applying two data augmentation techniques: Random Overlay and Random Convolution.}
\label{fig:data_augmentation}
\end{figure}

\textbf{Data Augmentations.}
\emph{Random Overlay} ~\cite{DBLP:journals/corr/abs-2011-13389} is an augmentation that performs linear interpolation between the original observation and another image randomly sampled from Places dataset ~\cite{DBLP:journals/pami/ZhouLKO018}, a dataset containing ten million scene photographs; 
\emph{Random Convolution} ~\cite{DBLP:conf/iclr/XuLYRN21} is an augmentation method that integrates a convolutional layer with random parameters into the observation, preserving the overall shape while adding a blurring effect to the texture features. 
The effects of the two data augmentations are shown in Figure~\ref{fig:data_augmentation}.
We adopt a combination of the above two data augmentations, that is, each batch randomly selects Random Overlay and Random Convolution with equal probability.

\subsection{DeepMind Control Suite}
\begin{figure}
  \centering
  \includegraphics[width=0.6\linewidth]{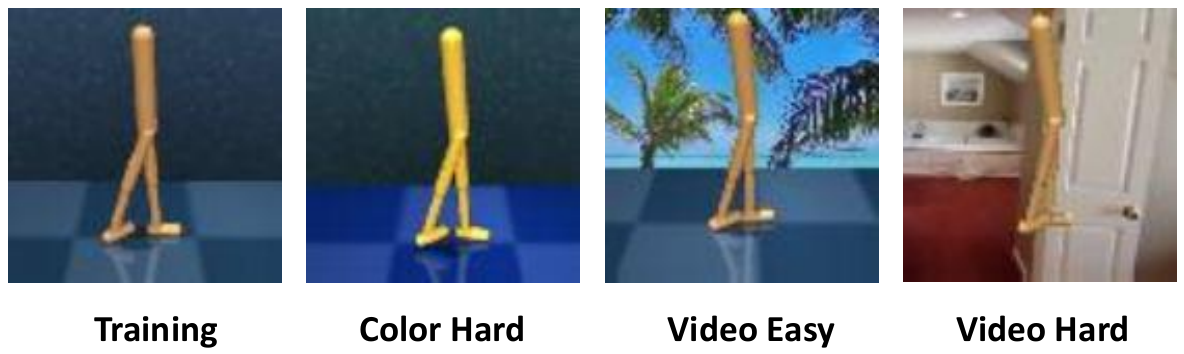}
  \caption{\textbf{Examples of DMControl-GB.}
    We separately present the training example observations and three test settings: ``Color Hard'', ``Video Easy'' and ``Video Hard''.}
\label{fig:DMC_env}
\end{figure}

We evaluate the generalization performance and sample efficiency of T2RD across five tasks in DMControl-GB: ``Walker Walk'', ``Walker Stand'', ``Ball in Cup Catch'', ``Finger Spin'', and ``Cartpole Swingup''. 
We conduct these assessments under three environmental settings: ``Color Hard'', ``Video Easy'', and ``Video Hard''.
In the ``Color Hard'' setting, the evaluation is carried out in environments where the colors of task-relevant objects (e.g., the agent’s body) and the background and ground are randomized. 
This setup directly tests the policy's ability to generalize to changes in the appearance of the agent itself, not just background variations.
In the ``Video'' setting, the evaluation is conducted in environments with various natural video backgrounds.
Figure~\ref{fig:DMC_env} provides examples of observations in both the training and three test environments. 
We train the agent for a total of 500,000 steps, conducting evaluations every 10,000 steps by calculating the average return over 30 episodes.
We stack three frames to form an observation with dimensions of 9 $\times$ 100 $\times$ 100. We then apply the random crop to the observation, resulting in dimensions of 9 $\times$ 84 $\times$ 84.

\begin{table*}
\caption{\textbf{Comparison with the SOTA on DMControl-GB.} We report the mean and standard deviation of episode returns for 5 different seeds. The best performance is indicated in bold. 
}
\centering
\resizebox{\textwidth}{!}{%
\begin{tabular} {l | c c c c c c c c c c}
    \toprule
    DMControl-GB & \multirow{2}{*}{DrQ ~\cite{DBLP:journals/corr/abs-2004-13649}} & \multirow{2}{*}{SODA ~\cite{DBLP:journals/corr/abs-2011-13389}}  & \multirow{2}{*}{SVEA ~\cite{DBLP:conf/nips/HansenSW21}} & \multirow{2}{*}{TLDA ~\cite{DBLP:conf/ijcai/YuanMMX00LX22}} & \multirow{2}{*}{PIE-G ~\cite{DBLP:conf/nips/YuanXYWWGX22}} & \multirow{2}{*}{SGQN ~\cite{DBLP:conf/nips/BertoinZZR22}} &  \multirow{2}{*}{BiT ~\cite{DBLP:journals/corr/abs-2312-01915}} & \multirow{2}{*}{TRP ~\cite{DBLP:conf/aaai/WangWHWLL24}} & \multirow{2}{*}{Visarl~\cite{DBLP:conf/iros/LiangTB24}} & \multirow{2}{*}{T2RD} \\
    (Color-Hard)& & & & & & & & & & \\
    
    \midrule
    Walker, Walk & 520 $\pm$ 91 & 692 $\pm$ 68 & 749 $\pm$ 61 & 823 $\pm$ 58 & \textbf{884 $\pm$ 20} & 780 $\pm$ 58 & 662 $\pm$ 64 & 823 $\pm$ 28 & 823 $\pm$ 55 & 849 $\pm$ 37 \\
    Walker, Stand & 770 $\pm$ 71 & 893 $\pm$ 12  & 933 $\pm$ 24 & 947 $\pm$ 26 & 960 $\pm$ 15 & 929 $\pm$ 12 & 903 $\pm$ 21 & 936 $\pm$ 13 & - & \textbf{964 $\pm$ 8} \\
    Ball\_in\_cup, Catch & 365 $\pm$ 210 & 949 $\pm$ 19 & 959 $\pm$ 5 & 930 $\pm$ 40 & 964 $\pm$ 7 & 864 $\pm$ 75 & 946 $\pm$ 25 & 826 $\pm$ 118 & 962 $\pm$ 14 & \textbf{965 $\pm$ 4} \\
    Finger, Spin & 776 $\pm$ 134 & 793 $\pm$ 128 & 972 $\pm$ 6 & 876 $\pm$ 45 & 922 $\pm$ 54 & 874 $\pm$ 31 & 889 $\pm$ 61 & 918 $\pm$ 34 & 823 $\pm$ 102 &  \textbf{984 $\pm$ 3} \\
    Cartpole, Swingup & 586 $\pm$ 52 & 805 $\pm$ 28 & 832 $\pm$ 23 & 760 $\pm$ 60 & 749 $\pm$ 46 & 777 $\pm$ 25 & 802 $\pm$ 32 & 799 $\pm$ 40 & \textbf{870 $\pm$ 21} &  857 $\pm$ 12 \\
    \midrule
    
    DMControl-GB & \multirow{2}{*}{DrQ} & \multirow{2}{*}{SODA} & \multirow{2}{*}{SVEA} & \multirow{2}{*}{TLDA} & \multirow{2}{*}{PIE-G} & \multirow{2}{*}{SGQN} &  \multirow{2}{*}{BiT} & \multirow{2}{*}{TRP}  & \multirow{2}{*}{Visarl} & \multirow{2}{*}{T2RD} \\
    (Video-Easy)& & & & & & & & & &\\
    
    \midrule
    Walker, Walk & 682 $\pm$ 89 & 768 $\pm$ 38  & 819 $\pm$ 71 & 868 $\pm$ 63 &  871 $\pm$ 22 & 865 $\pm$ 13 & 832 $\pm$ 42 & \textbf{871 $\pm$ 18} & 756 $\pm$ 42 &  821 $\pm$ 15 \\
    Walker, Stand & 873 $\pm$ 83 & 955 $\pm$ 13 & 961 $\pm$ 8 & 946 $\pm$ 6 & 957 $\pm$ 12 & 955 $\pm$ 9 & 960 $\pm$ 3 & 965 $\pm$ 8 & - & \textbf{972 $\pm$ 3} \\
    Ball\_in\_cup, Catch & 318 $\pm$ 157 & 857 $\pm$ 56 & 871 $\pm$ 106 & 855 $\pm$ 56 & 922 $\pm$ 20 &  \textbf{950 $\pm$ 24} & 899 $\pm$ 23 & 624 $\pm$ 244 &  802 $\pm$ 78 & 893 $\pm$ 43 \\
    Finger, Spin & 533 $\pm$ 119 & 695 $\pm$ 97  & 808 $\pm$ 33 & 756 $\pm$ 87 & 837 $\pm$ 107 & 956 $\pm$ 26 & 835 $\pm$ 25 & 816 $\pm$ 26  & 702 $\pm$ 83 & \textbf{966 $\pm$ 10} \\
    Cartpole, Swingup & 485 $\pm$ 105 & 758 $\pm$ 62  & 782 $\pm$ 27 & 671 $\pm$ 57 & 587 $\pm$ 61 & 761 $\pm$ 28 & 779 $\pm$ 34 & 733 $\pm$ 62 &  730 $\pm$ 32 & \textbf{847 $\pm$ 11} \\
    \midrule

    DMControl-GB & \multirow{2}{*}{DrQ} & \multirow{2}{*}{SODA} & \multirow{2}{*}{SVEA} & \multirow{2}{*}{TLDA} & \multirow{2}{*}{PIE-G} & \multirow{2}{*}{SGQN} & \multirow{2}{*}{BiT} &  \multirow{2}{*}{TRP} &  \multirow{2}{*}{Visarl} & \multirow{2}{*}{T2RD} \\
    (Video-Hard)& & & & & & & & & \\

    \midrule
    Walker, Walk & 121 $\pm$ 52 & 312 $\pm$ 32 & 385 $\pm$ 63 & 271 $\pm$ 55 & 600 $\pm$ 28 & \textbf{739 $\pm$ 21} & 325 $\pm$ 56 & 355 $\pm$ 69 & - & 444 $\pm$ 27 \\
    Walker, Stand & 252 $\pm$ 57 & 771 $\pm$ 83 & 834 $\pm$ 76 & 769 $\pm$ 59 & 852 $\pm$ 56 & 851 $\pm$ 24 & 829 $\pm$ 24 & 824 $\pm$ 2 & - &  \textbf{895 $\pm$ 43} \\
    Ball\_in\_cup, Catch & 100 $\pm$ 40 & 327 $\pm$ 100& 403 $\pm$ 174 & 457 $\pm$ 16 & \textbf{786 $\pm$ 47} & 782 $\pm$ 57 & 570 $\pm$ 71 & 274 $\pm$ 50 & - & 616 $\pm$ 38 \\
    Finger, Spin & 38 $\pm$ 13 & 302 $\pm$ 41 & 335 $\pm$ 58 & 224 $\pm$ 56 & 762 $\pm$ 59 & 822 $\pm$ 24 & 400 $\pm$ 13 & 261 $\pm$ 4 & - & \textbf{833 $\pm$ 18} \\
    Cartpole, Swingup & 136 $\pm$ 29 & 429 $\pm$ 64 & 393 $\pm$ 45 & 223 $\pm$ 62 & 401 $\pm$ 21 & 544 $\pm$ 43 & 526 $\pm$ 40 & 333 $\pm$ 27 & - &  \textbf{725 $\pm$ 37} \\

    \bottomrule
\end{tabular}%
}
\label{table:dmc}
\end{table*}

\textbf{T2RD exhibits the best generalization performance.} 
Table~\ref{table:dmc} shows the generalization results of T2RD and other SOTA algorithms in DMControl-GB tasks.
Our algorithm performs exceptionally well, achieving the best results in 9 out of 15 tasks. 
Particularly in the ``Cartpole Swingup'' task of the ``Video Hard'' setting, compared to the suboptimal algorithm, our method achieves a 33\% improvement.
Overall, T2RD demonstrates outstanding generalization performance across all tasks in various testing environments.

\begin{figure}
  \centering
  \includegraphics[width=\linewidth]{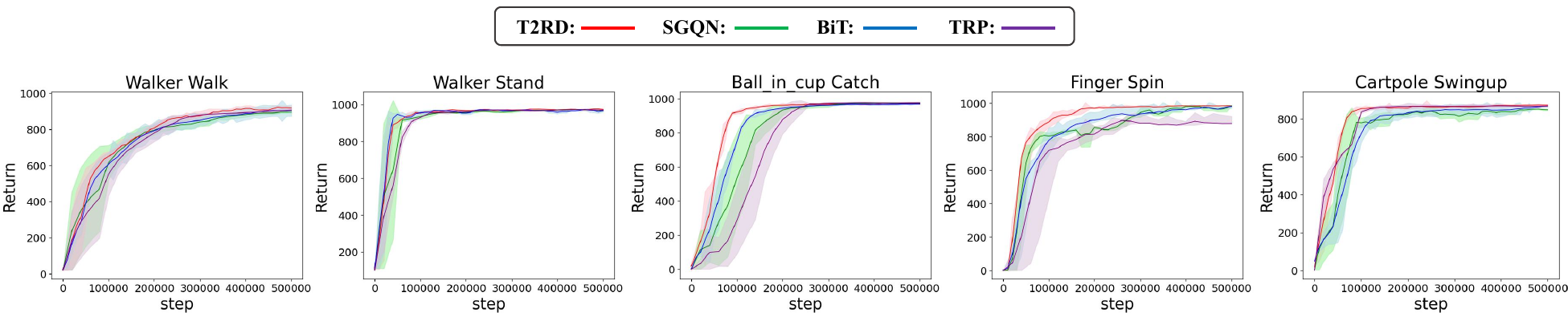}
  \caption{\textbf{Sample efficiency.} We visualize the training curves of T2RD, SGQN, BiT, and TRP. 
  The solid line represents the mean reported, and the shading represents the standard deviation.
  It can be seen that T2RD has the best sample efficiency and asymptotic performance.}
\label{sample_efficiency}
\end{figure}
\begin{figure}
  \centering
  \includegraphics[width=0.79\linewidth]{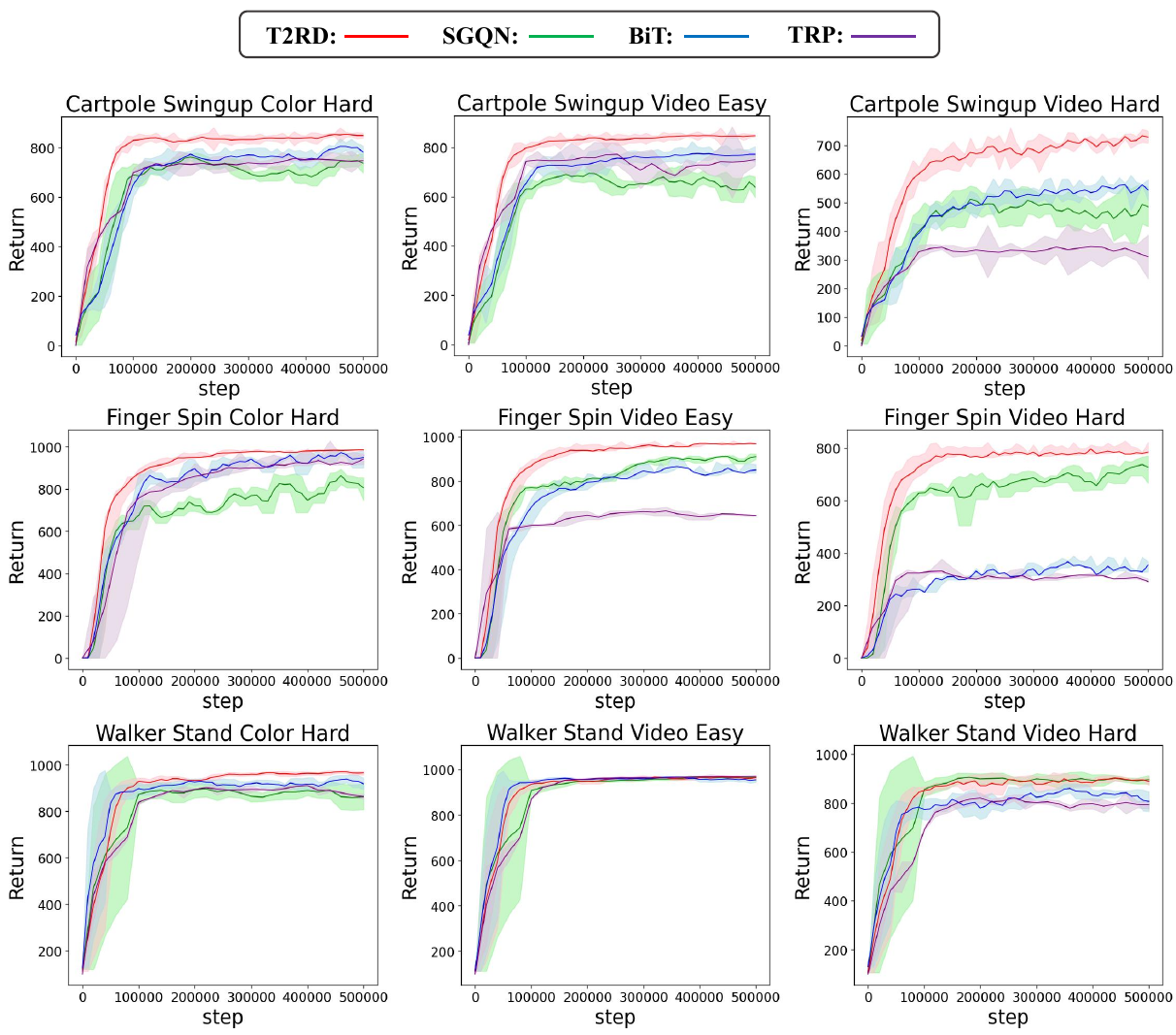}
  \caption{\textbf{Test curves on DMControl-GB.} 
  We compare the test curves of T2RD with SGQN, BiT, and TRP on three tasks in DMControl-GB: ``Cartpole Swingup'', ``Finger Spin'', and ``Walker Stand'', across three different environmental settings: Color Hard, Video Easy, and Video Hard.}
\label{fig:test_curve}
\end{figure}

\textbf{T2RD has the highest sample efficiency.}
In Figure~\ref{sample_efficiency}, we present the training curves of our method compared to SGQN, BiT, and TRP across five different tasks, with each algorithm running 5 seeds per task. 
Our method exhibits superior sample efficiency and asymptotic performance. Moreover, as illustrated in Figure ~\ref{fig:test_curve}, our method maintains excellent sample efficiency and asymptotic performance when generalized to previously unseen environments.

\begin{figure}
  \centering
  \includegraphics[width=\linewidth]{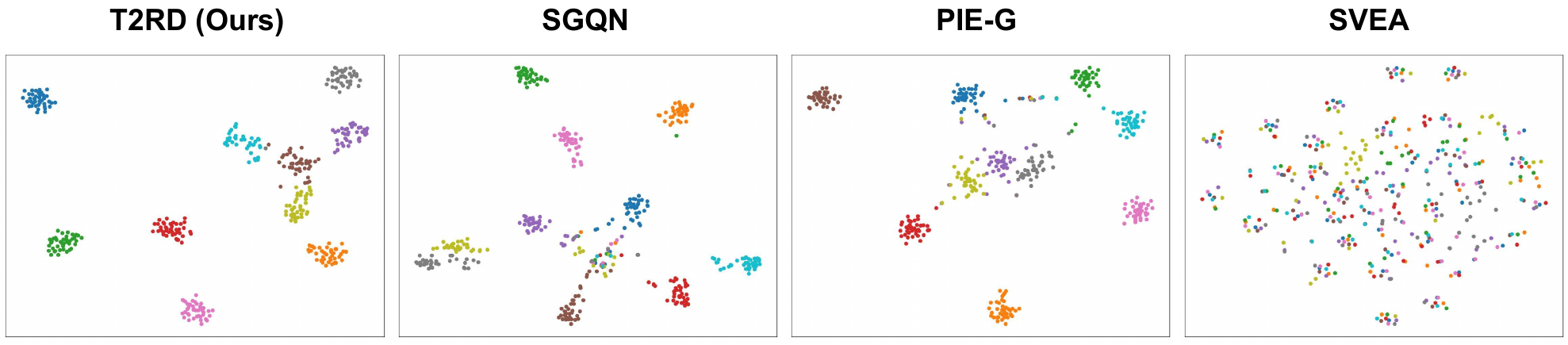}
  \caption{\textbf{t-SNE of representation.} We visualize the representations learned from T2RD, SGQN, PIEG, SVEA. T2RD exhibits the best intra-class cohesion and inter-class separation.}
\label{TSNE}
\end{figure}

\textbf{T2RD successfully learns task-relevant representations.}
In Figure ~\ref{TSNE}, we use t-SNE to visualize the features learned by our method and by SGQN, SVEA, and PIE-G.
We first randomly select 10 observations originating from different states in the ``Cartpole Swingup'' task and then replace their backgrounds with 40 unseen images. 
The 400 processed images are then embedded by the learned encoder of each algorithm. 
Observations from the same state are represented by the same color. 
From Figure ~\ref{TSNE}, we can observe that T2RD achieves the most effective aggregation for observations from the same state and effective separation between different states.

\begin{figure}
  \centering
  \includegraphics[width=0.75\linewidth]{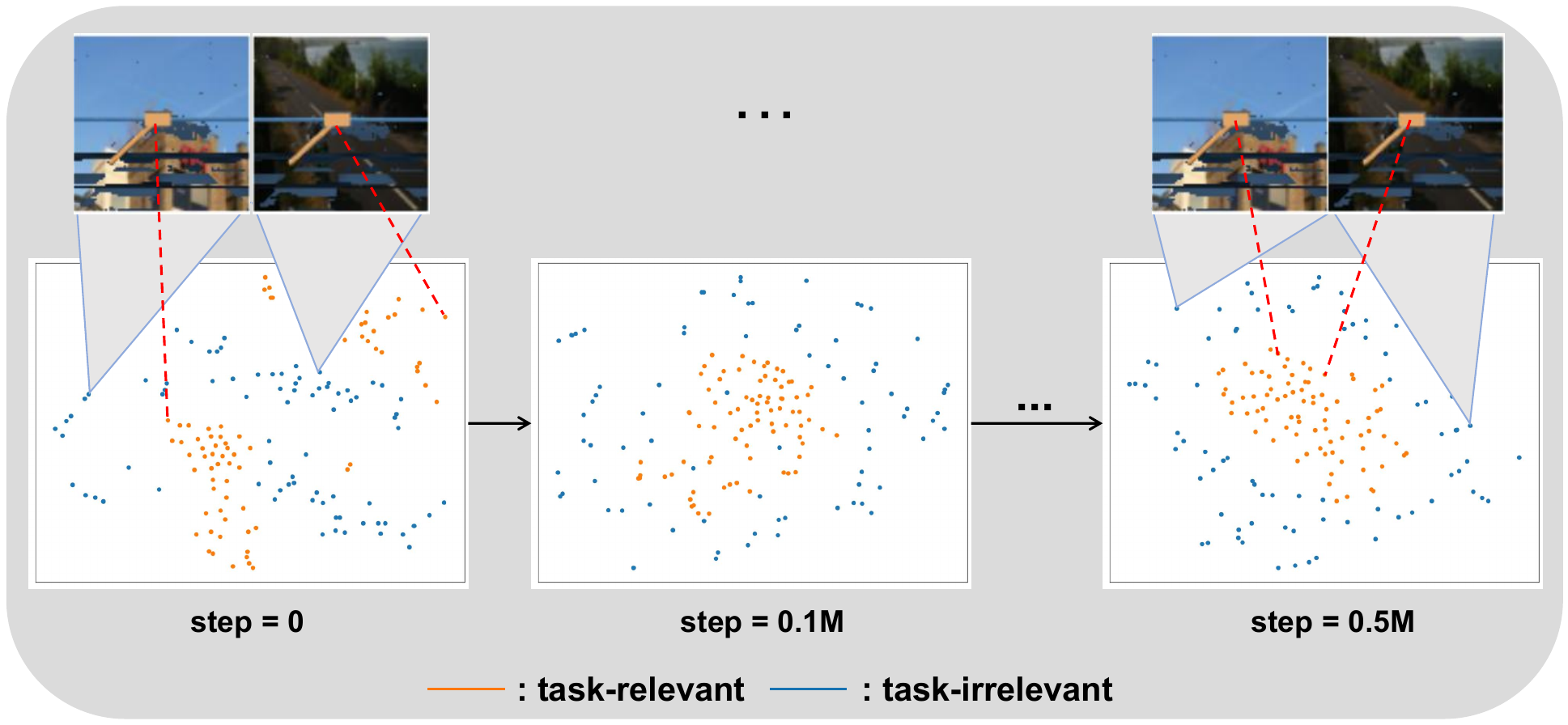}
  \caption{\textbf{t-SNE visualization of decoupled task-relevant and task-irrelevant features across training steps.} 
  }
\label{kenel_and_noise_TSNE}
\end{figure}

\textbf{T2RD can effectively achieve decoupling.} To validate that T2RD can effectively decouple task-relevant and task-irrelevant features, we conduct both quantitative and qualitative analyses.

First, we present t-SNE visualizations in Figure~\ref{kenel_and_noise_TSNE} to quantitatively analyze the learned representations over time. For this experiment, we take a single observation, create 80 variants by replacing its background with unseen images, and feed them into the T2RD encoder at different training checkpoints. As the figure illustrates, the task-relevant representations (orange dots) become progressively more concentrated into a cluster. This indicates that the model learns to map the same underlying state to a consistent representation, regardless of the visual background. Conversely, the task-irrelevant representations (blue dots) become more dispersed and gradually separable from the task-relevant features. This indicates that T2RD has learned the ability to decouple task-relevant and task-irrelevant features within visual observations.

\begin{figure}
  \centering
  \includegraphics[width=0.75\linewidth]{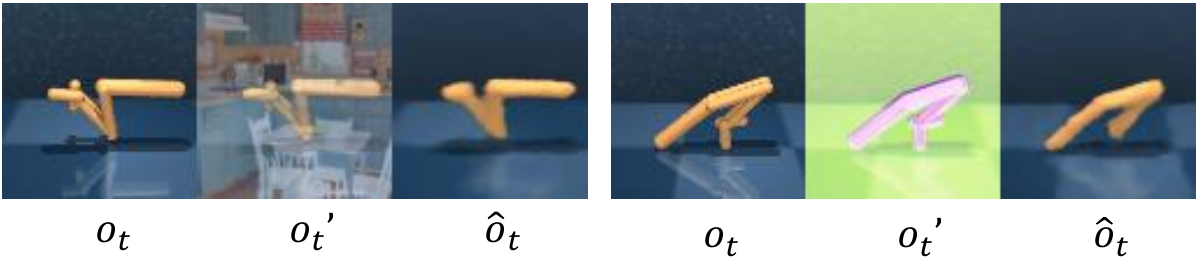}
  \caption{\textbf{Cross-reconstruction visualization.} We use the task-relevant features (content) $\mu_t'$ from the augmented observation $o_t'$ and the task-irrelevant features (style) $\epsilon_t$ from the original observation to perform cross-reconstruction, obtaining the reconstructed image $\hat{o}_t$. Successful reconstruction indicates that the T2RD learns truly decoupled representations.}
\label{fig:cross_reconstruction}
\end{figure}

Second, to provide more direct, qualitative proof, we conduct a cross-reconstruction experiment, visualized in Figure~\ref{fig:cross_reconstruction}. This involves reconstructing an original observation $o_t$ by combining the task-relevant representation $\mu_t'$ from a visually distinct \textit{augmented} view $o_t'$ with the task-irrelevant representation $\epsilon_t$ from the \textit{original} view $o_t$.
The result is a powerful demonstration of decoupling: the reconstructed image successfully captures the agent's posture (task-relevant) from the original observation, while completely ignoring the distracting background or color perturbation (task-irrelevant) introduced in the augmented view. 
Together, these visualizations provide strong evidence that T2RD effectively decouples task-relevant content from task-irrelevant style.


\begin{figure}
  \centering
  \includegraphics[width=0.8\linewidth]{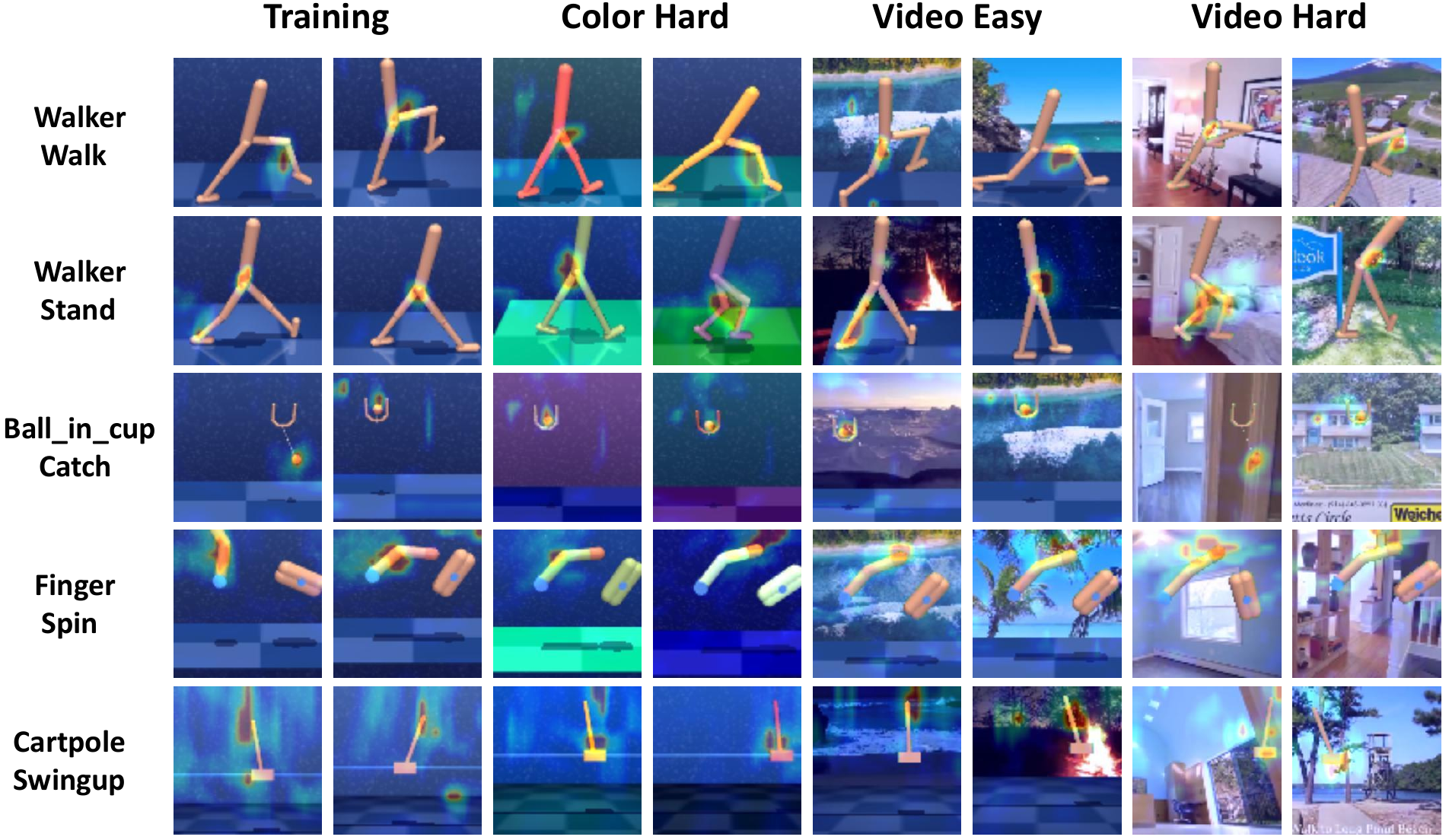}
  \caption{\textbf{Visualizations of attention maps.} We visualize the attention areas of our algorithm under different backgrounds in DMControl-GB. }
\label{HeatMaps}
\end{figure}

\textbf{T2RD accurately identifies task-relevant regions.}
In Figure~\ref{HeatMaps}, we visualize the attention regions of T2RD under different backgrounds in DMControl-GB: ``Training'', ``Color Hard'', ``Video Easy'', and ``Video Hard''.
T2RD can accurately perceive task-relevant regions in visual observations, regardless of the presence of background interference. 
This provides the agent with reliable task-relevant representations for learning generalizable policies.

\subsection{Robotic Manipulation}

\begin{figure}
  \centering
  \includegraphics[width=0.65\linewidth]{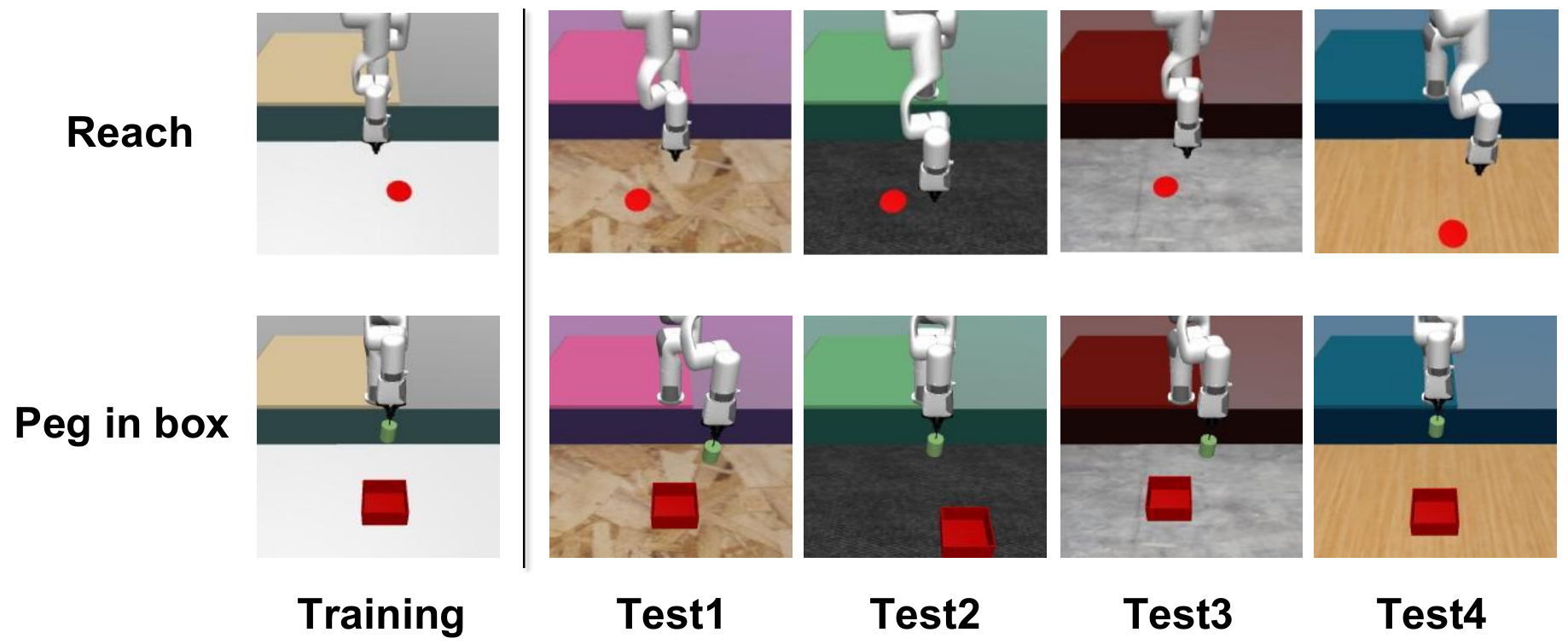}
  \caption{\textbf{Examples of Robotic Manipulation.}
   We present training example observations and four test settings for the ``Reach'' and ``Peg in box'' tasks.}
\label{fig:robot_mani_env}
\end{figure}

In addition to DeepMind control tasks, we also conduct experiments on Robotic Manipulation tasks.
We consider two goal-reaching tasks: 1) \textit{Reach}, where the robotic arm attempts to reach a red mask on the tabletop; 2) \textit{Peg in box}, where the objective is to control the robotic arm to insert a peg into a box. 
The dimensions of the observation are 3 $\times$ 100 $\times$ 100. We perform a random crop on the observation to obtain a cropped observation with dimensions of 3 $\times$ 84 $\times$ 84. 
The agent is trained in a fixed background and tested in environments where key information is retained while the texture or color of the background is disturbed.
We train the agent for 500,000 steps, conducting tests every 10,000 steps by calculating the average episode return over 30 episodes.
For each task, we employ four test environments: Test1, Test2, Test3, and Test4. 
Figure~\ref{fig:robot_mani_env} provides examples of observations in both the training and four test environments. 
We compare our algorithm against SAC, SODA, SVEA, SGQN, BiT, and TRP.

The results are shown in Table \ref{table:robot}, where T2RD achieves state-of-the-art performance in 7 out of 8 test environments. 
In the Reach task, T2RD shows the best performance in Test1, Test3, and Test4 environments, and even in Test2, we are the second-best algorithm.
Compared to the second-best algorithm, our average performance improves by 40.1\%.
In the Peg in box task, T2RD outperforms all other algorithms in all test environments.
Compared to the second-best algorithm, our average performance improves by 57.1\%.
The above experimental results demonstrate the superiority of our algorithm. 
Additionally, T2RD proves applicable to various VRL tasks.

\begin{table*}
\setlength\tabcolsep{3pt}
\caption{\textbf{Comparison with the SOTA on Robotic Manipulation tasks.} We report the mean and standard deviation of episode returns for 5 different seeds. The best performance is indicated in bold.}
\centering
\resizebox{\textwidth}{!}{%
\begin{tabular}{l | c | c c c c c c c}
    \toprule
    Task & Environment & SAC ~\cite{DBLP:conf/icml/HaarnojaZAL18} & SODA ~\cite{DBLP:journals/corr/abs-2011-13389} & SVEA ~\cite{DBLP:conf/nips/HansenSW21} & SGQN ~\cite{DBLP:conf/nips/BertoinZZR22} & BiT ~\cite{DBLP:journals/corr/abs-2312-01915} & TRP ~\cite{DBLP:conf/aaai/WangWHWLL24} & T2RD \\
    \midrule
    \multirow{5}{*}{Reach} & Test1 & $-$20.9 $\pm$ 16 & $-$30.9 $\pm$ 43 & $-$17.6 $\pm$ 10 & 14.4 $\pm$ 14 & $-$9 $\pm$ 18 & 18.0 $\pm$ 7 & \textbf{25.1 $\pm$ 8} \\
     & Test2 & $-$21.9 $\pm$ 14 & $-$20.2 $\pm$ 29 & $-$2.1 $\pm$ 39 & \textbf{31.0 $\pm$ 3} & $-$3 $\pm$ 21 & $-$25.4 $\pm$ 19 & 13.2 $\pm$ 10 \\
     & Test3 & $-$43.2 $\pm$ 6 & $-$68.4 $\pm$ 30 & 1.4 $\pm$ 29 & 29.2 $\pm$ 7 & $-$28 $\pm$ 30 & 29.6 $\pm$ 2 & \textbf{29.8 $\pm$ 2} \\
     & Test4 & $-$36.7 $\pm$ 3 & $-$37.9 $\pm$ 26 & $-$22.2 $\pm$ 24 & $-$5.6 $\pm$ 20 & $-$27 $\pm$ 34 & 24.1 $\pm$ 5  & \textbf{28.3 $\pm$ 8} \\
     & Avg & $-$30.6 & $-$39.3 & $-$10.1 & 17.2 & $-$16.7 & 11.5 & \textbf{24.1} \\
    \midrule
    \multirow{5}{*}{Peg in box} & Test1 & $-$59.6 $\pm$ 26 & 16.9 $\pm$ 44 & $-$21.3 $\pm$ 10 & $-$72.1 $\pm$ 14 & $-$16 $\pm$ 57 & 63.5 $\pm$ 17 & \textbf{164.1 $\pm$ 34} \\
     & Test2 & $-$60.5 $\pm$ 12 & 0.7 $\pm$ 30 & 96.8 $\pm$ 41 & 110.7 $\pm$ 3 & 27 $\pm$ 43 & 55.6 $\pm$ 31 & \textbf{129.4 $\pm$ 91} \\
     & Test3 & $-$48.8 $\pm$ 17 & 73.6 $\pm$ 31 & 40.5 $\pm$ 28 & 154.6 $\pm$ 7 & 55 $\pm$ 52 & 162.5 $\pm$ 23 & \textbf{183.6 $\pm$ 24} \\
     & Test4 & $-$89.6 $\pm$ 31 & $-$16 $\pm$ 83 & 33.2 $\pm$ 43 & $-$70.9 $\pm$ 41 & 72 $\pm$ 48 & 117.4 $\pm$ 47 & \textbf{149.8 $\pm$ 82} \\
     & Avg & $-$64.6 & 18.8 & 37.3 & 30.5  & 34.5 & 99.7 & \textbf{156.7} \\
    \bottomrule
\end{tabular}%
}
\label{table:robot}
\end{table*}
\begin{figure}
  \centering
  \includegraphics[width=0.61\linewidth]{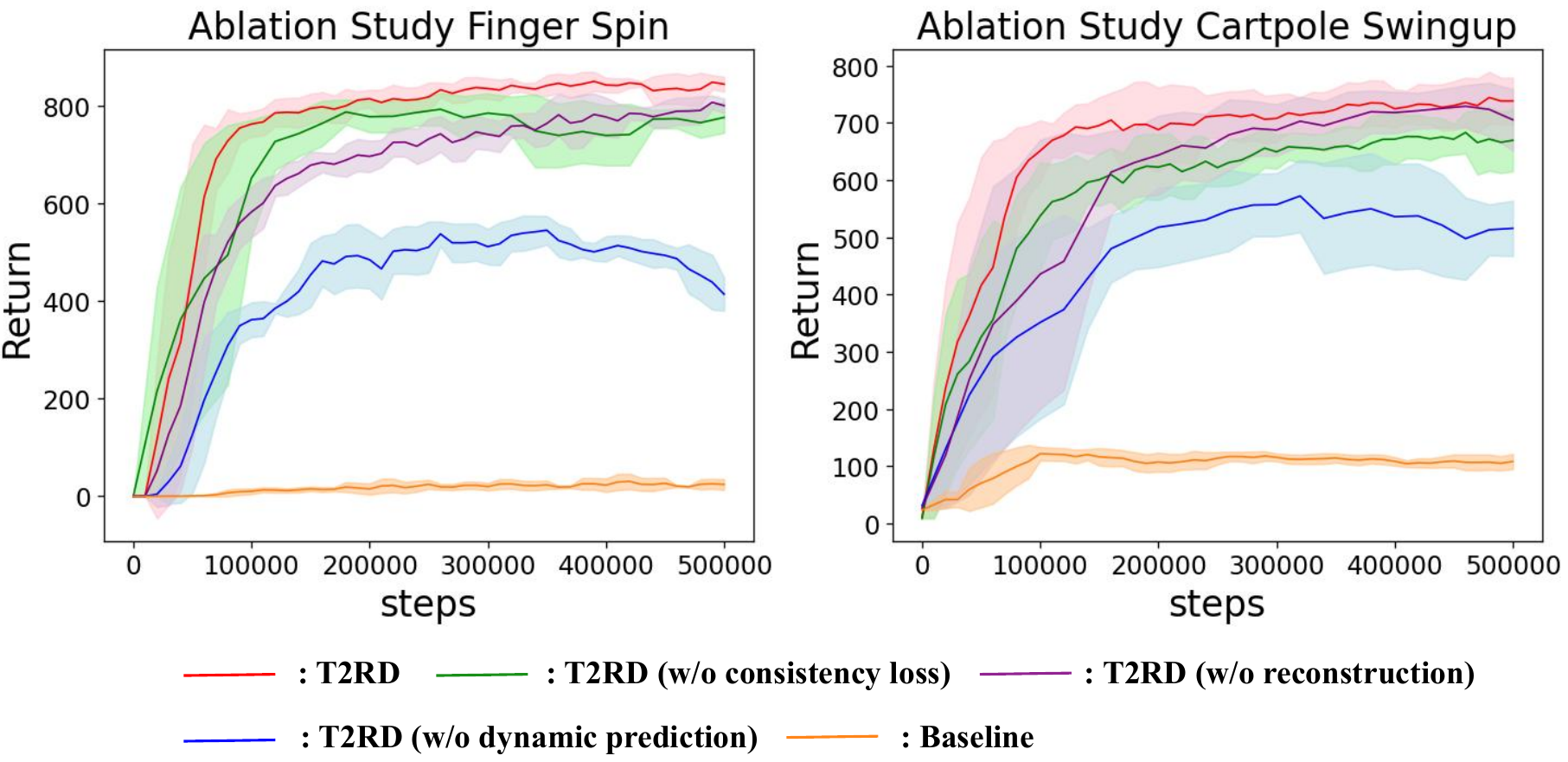}
  \caption{\textbf{Ablation Study on DMControl-GB.} We conduct ablation studies on the ``Cartpole Swingup'' and ``Finger Spin'' tasks under the ``Video Hard'' setting.}
\label{Ablation_DMC}
\end{figure}

\begin{table}
\setlength\tabcolsep{2pt}
\caption{\textbf{Ablation study on Robotic Manipulation tasks.} In which ``cons.'', ``recon.'', and ``dyn.'' respectively represent the ``task-relevant representation consistency'', ``cross-reconstruction'', and ``cross-dynamic prediction'' modules. 
}
\centering
\resizebox{0.75\textwidth}{!}{%
\begin{tabular}{c | c c c | c c c c}
    \toprule
    \multirow{1}{*}{Robotic Manipulation} & \multirow{1}{*}{cons.} & \multirow{1}{*}{recon.} & \multirow{1}{*}{dyn.} & Test1 & Test2 & Test3 & Test4 \\
    
    \midrule
   \multirow{4}{*}{Reach} & \checkmark & & \checkmark & 24.1 $\pm$ 12 & 9.5 $\pm$ 12 & 20.8 $\pm$ 9 & 24.8 $\pm$ 12  \\
    & & \checkmark & \checkmark  & 27.6 $\pm$ 6 & 8.3 $\pm$ 25 & 21.0 $\pm$ 9 & 27.7 $\pm$ 6.1  \\
     & \checkmark & \checkmark & & \textbf{27.7 $\pm$ 5} & 10.1 $\pm$ 17.3 &25.2 $\pm$ 8 & 27.4 $\pm$ 28 \\
    & \checkmark & \checkmark & \checkmark & 25.1 $\pm$ 8 & \textbf{13.2 $\pm$ 10} & \textbf{29.8 $\pm$ 2} & \textbf{28.3 $\pm$ 8} \\
    \toprule

    \multirow{4}{*}{Peg in box} & \checkmark & & \checkmark & 136 $\pm$ 62 & 106 $\pm$ 66 & 164 $\pm$ 25 & 135 $\pm$ 45  \\
    & & \checkmark & \checkmark  & 146 $\pm$ 24 & 68.5 $\pm$ 115 & 136 $\pm$ 58 & 129 $\pm$ 57  \\
     & \checkmark & \checkmark & & 151 $\pm$ 38 & 118 $\pm$ 36 & 171 $\pm$ 21 & 142 $\pm$ 16 \\
    & \checkmark & \checkmark & \checkmark & \textbf{164 $\pm$ 34} & \textbf{129 $\pm$ 91} & \textbf{183 $\pm$ 24} & \textbf{149 $\pm$ 82} \\
    
    \bottomrule
\end{tabular}%
 }
\label{table:Robotic ablation}
\end{table}

\subsection{Ablation Study}
We conduct ablation experiments on the DMControl-GB to assess the impact of each objective function proposed by T2RD.
The experimental results are shown in Figure~\ref{Ablation_DMC}, where ``T2RD (w/o consistency loss)'' marks the absence of the task-relevant representation consistency module; 
``T2RD (w/o reconstruction)'' indicates the exclusion of the cross-reconstruction module; 
``T2RD (w/o dynamic prediction)'' denotes the omission of the cross-dynamic prediction module; 
``Baseline'' represents the fundamental version of the T2RD algorithm, which omits these three objective functions but retains the combined data augmentation.


From Figure ~\ref{Ablation_DMC}, it's clear that T2RD achieves superior sample efficiency and asymptotic performance.
The comparison between ``T2RD (w/o dynamic prediction)'' and T2RD underscores the critical role of dynamic prediction in learning task-relevant representations within our decoupling framework.
Nonetheless, even without dynamic prediction, methods that separate content and style at the visual level remain effective, continuing to surpass other SOTA algorithms.
Additionally, the ``T2RD (w/o reconstruction)'' curve shows that the decoupling of content and style facilitates the learning of representations that are robust to visual changes.
Finally, ``T2RD (w/o consistency loss)'' also achieves excellent experimental results, but there is still a certain gap compared to T2RD.
This suggests that the task-relevant representation consistency module contributes to learning domain-invariant representations, which promotes generalization.

Additionally, we conduct ablation experiments on three modules in Robotic Manipulation tasks, with the results shown in Table~\ref{table:Robotic ablation}. 
The experimental results show that removing any one of the three modules leads to a decrease in performance. 
In summary, these three modules are complementary and work together to achieve the decoupling of task-relevant and task-irrelevant features, thereby promoting the generalization of VRL.



\begin{table}
\setlength\tabcolsep{2pt}
\caption{\textbf{Variant experiments on DMControl-GB.} We select two tasks: ``Cartpole Swingup'' and ``Finger Spin'', to conduct variant experiments in the ``Color Hard'', ``Video Easy'' and ``Video Hard'' setting.}
\centering
\resizebox{0.55\textwidth}{!}{%
\begin{tabular}{l | c | c cc}
    \toprule
    Task & Variation & Color Hard & Video Easy &  Video Hard \\
    \midrule
     \multirow{5}{*}{\shortstack{Finger, \\ Spin}}
     & T2RD-Alt. Branch & 965 $\pm$ 17 & 897 $\pm$ 80  & 665 $\pm$ 79 \\
     & T2RD-Dual & 976 $\pm$ 5 & 932 $\pm$ 14  & 655 $\pm$ 37 \\
     & T2RD-NonCross-Recon. & 913 $\pm$ 73 & 920 $\pm$ 39  & 676 $\pm$ 68 \\
     & T2RD-NonCross-Pred. & 934 $\pm$ 46 & 888 $\pm$ 61  & 609 $\pm$ 49 \\
     & T2RD (w/o EMA) & 959 $\pm$ 23 & 922 $\pm$ 67  & 720 $\pm$ 72 \\
     & T2RD-Pixel Recon. & 888 $\pm$ 83 & 803 $\pm$ 76  & 486 $\pm$ 75 \\
     & T2RD & \textbf{984 $\pm$ 3} & \textbf{966 $\pm$ 10}  & \textbf{833 $\pm$ 18} \\
    \midrule

     \multirow{5}{*}{\shortstack{Cartpole, \\ Swingup}} 
     & T2RD-Alt. Branch & 826 $\pm$ 28 & 824 $\pm$ 19  & 614 $\pm$ 72 \\
     & T2RD-Dual & 838 $\pm$ 23 & 831 $\pm$ 37  & 679 $\pm$ 56 \\
     & T2RD-NonCross-Recon. & 844 $\pm$ 27 & 838 $\pm$ 8  & 659 $\pm$ 77 \\
     & T2RD-NonCross-Pred. & 799 $\pm$ 33 & 748 $\pm$ 44  & 519 $\pm$ 68 \\
     & T2RD (w/o EMA) & 824 $\pm$ 23 & 828 $\pm$ 13  & 680 $\pm$ 64 \\
     & T2RD-Pixel Recon. & 800 $\pm$ 63 & 786 $\pm$ 70  & 653 $\pm$ 87 \\
     & T2RD & \textbf{857 $\pm$ 12} & \textbf{847 $\pm$ 11}  & \textbf{725 $\pm$ 37} \\
    \bottomrule
\end{tabular}%
}
\label{table:variation}
\end{table}

\subsection{Variant Experiments}
To provide deeper insights into our key design choices beyond the module-level ablations in Section 5.3, we conduct a series of more fine-grained variant experiments. The results are shown in Table~\ref{table:variation} and Table~\ref{tab varient augmentation results}. We analyze structural variants, key mechanisms, and data augmentation methods.

\subsubsection{Analysis of Structural Variants}

We first explore several structural variants of T2RD. ``T2RD-Alt. Branch'' employs an alternative reconstruction branch using $\mu_{t}$ and $\epsilon_{t}'$ to reconstruct $z_{t}'$. 
In Table~\ref{table:variation}, ``T2RD-Dual'' indicates the simultaneous use of both the primary and alternative reconstruction branches. ``T2RD-Pixel Recon.'' denotes pixel-based reconstruction, meaning that instead of reconstructing back to the latent representation $z_{t}$, we reconstruct back to the observation $o_{t}$.

The experimental results in Table~\ref{table:variation} reveal that both the ``T2RD-Alt. Branch'' and ``T2RD-Dual'' are less effective than our primary T2RD design. We argue that the task-irrelevant representations of the augmented observations ($\epsilon_t'$) contain too many distracting elements, leading to algorithm divergence. Furthermore, we find that reconstruction based on latent representations is more effective than pixel-based reconstruction (``T2RD-Pixel Recon.'' in Table~\ref{table:variation}), a finding also supported by previous research~\cite{DBLP:conf/icra/OkadaT21}.

\subsubsection{Analysis of Key Design Mechanisms}

To provide a more fine-grained analysis, we evaluate the specific impact of our cross-domain strategy and the EMA update mechanism.

\textbf{Cross-Domain vs. Same-Domain.} To isolate the effect of our cross-domain strategy, we test two variants. ``T2RD-NonCross-Recon.'' performs same-domain reconstruction (using $\mu_{t}$ and $\epsilon_{t}$ to reconstruct $z_{t}$). ``T2RD-NonCross-Pred.'' performs same-domain prediction (using the augmented representation $\mu_t'$ to predict the next augmented representation $\mu_{t+1}'$). As shown in Table~\ref{table:variation}, T2RD with its cross-domain objectives significantly outperforms both of these variants. This demonstrates that enforcing consistency across different visual domains (original vs. augmented) is a powerful constraint that promotes the learning of more robust and generalizable representations.

\textbf{Importance of EMA.} We introduce ``T2RD (w/o EMA)'', where the target network parameters are updated via a direct copy from the online network instead of the EMA update. The results in Table~\ref{table:variation} show a noticeable drop in performance compared to the standard T2RD. This empirically confirms that the EMA is a crucial component for stabilizing the self-supervised training process by providing a consistent and slowly evolving target, which helps prevent representational collapse.

\subsubsection{Analysis of Data Augmentation Methods}

In addition to structural variant experiments, we conduct variant experiments on data augmentation, comparing the experimental results of the T2RD algorithm using Random Overlay, Random Convolution, and combined data augmentation. The experimental results are shown in Table \ref{tab varient augmentation results}, where T2RD (conv) indicates the use of Random Convolution as a replacement for the combined data augmentation technique in the T2RD algorithm. Similarly, T2RD (overlay) refers to the use of Random Overlay data augmentation. The experimental results demonstrate that combined data augmentation significantly improves the generalization performance of the algorithm compared to single data augmentation methods. As discussed in the methods section, T2RD greatly benefits from combined data augmentation due to its strong decoupling ability. T2RD can fully exploit the advantages brought by increased augmentation diversity, alleviating the issue of generalization bias. Additionally, thanks to its strong decoupling capabilities, T2RD is less affected by conflicts between different augmentations, thus maximizing the benefits of combined data augmentation.

\begin{table}
\caption{\textbf{Experimental results of T2RD using Random Overlay, Random Convolution, and our combined data augmentation approaches.}}
  \label{tab:data augmentation results}
  \centering
  \resizebox{0.5\textwidth}{!}{%
  \begin{tabular}{l | c | c cc}
    \toprule
    Task & Data Augmentation & Color Hard & Video Easy &  Video Hard \\
    \midrule
    \multirow{3}{*}{\shortstack{Finger, \\ Spin}}
    & T2RD (conv) & $973 \pm 8$ & $605 \pm 19$ & $227 \pm 35$ \\
    & T2RD (overlay) & $882 \pm 23$ & $866 \pm 8$ & $464 \pm 62$ \\
    & T2RD & \textbf{984 $\pm$ 3} & \textbf{966 $\pm$ 10}& \textbf{833 $\pm$ 18}\\
    \midrule
    \multirow{3}{*}{\shortstack{Cartpole, \\ Swingup}}
    & T2RD (conv) & $ 854 \pm 11 $ & $ 589 \pm 33 $ & $ 213 \pm 36 $ \\
    & T2RD (overlay)  & $ 751 \pm 47 $ & $ 749 \pm 52 $ & $ 519 \pm 42 $ \\
    & T2RD & \textbf{857 $\pm$ 12} & \textbf{847 $\pm$ 11} & \textbf{725 $\pm$ 37} \\
    \midrule
    \multirow{3}{*}{\shortstack{Walker, \\ Stand}}
    & T2RD (conv) & $ 948 \pm 7 $ & $ 709 \pm 21 $ & $ 413 \pm 46 $ \\
    & T2RD (overlay)  & $ 902 \pm 29 $ & $ 964 \pm 6 $ & $ 845 \pm 22 $ \\
    & T2RD & \textbf{964 $\pm$ 8} & \textbf{972 $\pm$ 3} & \textbf{895 $\pm$ 43} \\
    \bottomrule
  \end{tabular}
  }
\label{tab varient augmentation results}
\end{table}

\begin{table}
\caption{\textbf{SODA and SVEA with Specific Data Augmentations.} Results for SODA and SVEA trained with single augmentations (Random Convolution (Conv) or Random Overlay (Overlay)) on ``Color-Hard'' (color perturbations) and ``Video-Easy'' settings (background perturbations).}
\centering
\resizebox{0.6\textwidth}{!}{%
\begin{tabular} {l | c c | c c }
    \toprule
    DMControl-GB & \multirow{2}{*}{SODA (Conv)} & \multirow{2}{*}{SODA (Overlay)} & \multirow{2}{*}{SVEA (Conv)} & \multirow{2}{*}{SVEA (Overlay)}\\
    (\textbf{Color-Hard})& & & & \\
    
    \midrule
    Walker, Walk & \textbf{697 $\pm$ 66} & 692 $\pm$ 68 & \textbf{760 $\pm$ 145} & 749 $\pm$ 61 \\
    Walker, Stand & \textbf{930 $\pm$ 12} & 893 $\pm$ 12  & \textbf{942 $\pm$ 26} & 933 $\pm$ 24 \\
    Ball\_in\_cup, Catch & 892 $\pm$ 37 & \textbf{949 $\pm$ 19} & \textbf{961 $\pm$ 7} & 959 $\pm$ 5 \\
    Finger, Spin & \textbf{892 $\pm$ 52} & 793 $\pm$ 128 & \textbf{977 $\pm$ 5} & 972 $\pm$ 6 \\
    Cartpole, Swingup & \textbf{831 $\pm$ 21} & 805 $\pm$ 28 & \textbf{837 $\pm$ 23} & 832 $\pm$ 23   \\
    \midrule
    
    DMControl-GB & \multirow{2}{*}{SODA (Conv)} & \multirow{2}{*}{SODA (Overlay)} & \multirow{2}{*}{SVEA (Conv)} & \multirow{2}{*}{SVEA (Overlay)} \\
    (\textbf{Video-Easy})& & & & \\
    
    \midrule
    Walker, Walk & 635 $\pm$ 48 & \textbf{768 $\pm$ 38}  & 612 $\pm$ 144 & \textbf{819 $\pm$ 71} \\
    Walker, Stand & 903 $\pm$ 56 & \textbf{955 $\pm$ 13} & 795 $\pm$ 70& \textbf{961 $\pm$ 8} \\
    Ball\_in\_cup, Catch & 539 $\pm$ 111 & \textbf{875 $\pm$ 56} & 659 $\pm$ 110 & \textbf{871 $\pm$ 106} \\
    Finger, Spin & 363 $\pm$ 185 & \textbf{695 $\pm$ 97}  & 764 $\pm$ 86 & \textbf{808 $\pm$ 33} \\
    Cartpole, Swingup & 474 $\pm$ 143 & \textbf{758 $\pm$ 62}  & 606 $\pm$ 86 & \textbf{782 $\pm$ 27} \\
    
    \bottomrule
\end{tabular}%
}
\label{table:bias}
\end{table}

\subsection{Analysis of Combined Data Augmentation and Generalization Bias}

Many VRL algorithms exhibit \textit{generalization bias} by relying on a single data augmentation, where performance becomes overly dependent on the specific transformations seen during training.
For instance, an agent trained only on ``color'' augmentation (like Random Convolution) is good at handling color changes but poor at handling background changes, and vice versa.

To empirically demonstrate this, we conduct an experiment where we train SODA~\cite{DBLP:journals/corr/abs-2011-13389} and SVEA~\cite{DBLP:conf/nips/HansenSW21} with each of these single augmentations separately. 
The results, presented in Table~\ref{table:bias}, clearly confirm the existence of this bias. 
In the ``Color-Hard'' setting, agents trained with Random Convolution generally outperform those trained with Random Overlay. 
Conversely, in the ``Video-Easy'' setting, agents trained with Random Overlay are significantly superior. 
This confirms that relying on a single augmentation leads to policies that are brittle to other forms of visual change.


Our proposed combined data augmentation method effectively mitigates this bias.
In Table~\ref{tab:combined_data_augmentation}, we apply the combined augmentation strategy to several baseline algorithms.
As shown, the performance of baselines such as SODA~\cite{DBLP:journals/corr/abs-2011-13389} and SVEA~\cite{DBLP:conf/nips/HansenSW21} not only improves significantly but also becomes more balanced across the three distinct test environments (``Color Hard'', ``Video Easy'', and ``Video Hard'').
This provides strong evidence that our combined strategy alleviates generalization bias by breaking the agent’s dependence on a single type of visual perturbation.

However, this analysis also highlights the unique synergy between our architecture and the augmentation strategy.
While the baselines benefit from the combined augmentation, T2RD’s performance (as shown in Table~\ref{tab:combined_data_augmentation}) remains markedly superior.
We argue that although increasing augmentation diversity is beneficial, it can also introduce interference or conflicts among different transformations.
Thanks to T2RD’s robust decoupling capability, it can fully exploit this increased diversity—maximizing the advantages of combined data augmentation where other methods may struggle.
For example, the slight performance drop observed for SGQN~\cite{DBLP:conf/nips/BertoinZZR22} on one task suggests that, without a strong decoupling mechanism, increased augmentation diversity may sometimes interfere with methods that rely heavily on fine-grained visual cues such as saliency maps.
This demonstrates that the observed gains arise not from augmentation alone, but from the powerful combination of our novel architecture and the augmentation strategy.

Finally, we explore the impact of the sampling probability \( p \) for Random Overlay within our combined approach. 
The specific experimental results are shown in Table~\ref{table:sensitive}. The results indicate that when \( p = 0.5 \), i.e., when Random Convolution and Random Overlay are chosen with equal probability, we observe the best experimental outcomes, reinforcing the idea that balanced diversity is key to effective data augmentation.

\begin{table}
\caption{\textbf{The universal applicability of combined data augmentation.} We demonstrate changes in experimental results for other algorithms after applying the same combined data augmentation as T2RD.}
\centering
\resizebox{0.7\textwidth}{!}{%
\begin{tabular}{l | c | c | c|c}
\toprule
    Task & Algorithm & Color-Hard & Video-Easy & Video-Hard \\
\midrule
\multirow{6}{*}{\shortstack{Finger, \\ Spin}} 
& SODA & $793 \pm 128 \rightarrow 927 \pm 27$ & $695 \pm 97 \rightarrow 847 \pm 63$ & $302 \pm 41 \rightarrow 522 \pm 72$ \\
& SVEA & $972 \pm 6 \rightarrow 977 \pm 5$ & $808 \pm 33 \rightarrow 904 \pm 22$ & $335 \pm 58 \rightarrow 650 \pm 42$ \\
& TLDA & $876 \pm 45 \rightarrow 881 \pm 18$ & $756 \pm 87 \rightarrow 860 \pm 26$ & $224 \pm 56 \rightarrow 450 \pm 47$ \\
& SGQN & $874 \pm 31 \rightarrow 923 \pm 41$ & $956 \pm 26 \rightarrow 906 \pm 16$ & $822 \pm 24 \rightarrow 746 \pm 27$ \\
& BiT & $889 \pm 61 \rightarrow 910 \pm 52$ & $835 \pm 25 \rightarrow 820 \pm 27$ & $400 \pm 13 \rightarrow 507 \pm 47$ \\
& T2RD & $\mathbf{984 \pm 3}$ & $\mathbf{966 \pm 10}$ & $\mathbf{833 \pm 18}$ \\
\midrule
\multirow{6}{*}{\shortstack{Cartpole, \\ Swingup}} 
& SODA & $805 \pm 28 \rightarrow 833 \pm 23
$ & $758 \pm 62 \rightarrow 777 \pm 37$ & $429 \pm 64 \rightarrow 481 \pm 45$ \\
& SVEA & $832 \pm 23 \rightarrow 842 \pm 27$ & $782 \pm 27 \rightarrow 828 \pm 32$ & $393 \pm 45 \rightarrow 592 \pm 123$ \\
& TLDA & $760 \pm 60 \rightarrow 786 \pm 53$ & $671 \pm 57 \rightarrow 705 \pm 56$ & $223 \pm 62 \rightarrow 329 \pm 47$ \\
& SGQN & $777 \pm 25 \rightarrow 800 \pm 26$ & $761 \pm 28 \rightarrow 758 \pm 35$ & $544 \pm 43 \rightarrow 596 \pm 92$ \\
& BiT & $802 \pm 32 \rightarrow 838 \pm 24$ & $779 \pm 34 \rightarrow 810 \pm 20$ & $526 \pm 40 \rightarrow 585 \pm 41$ \\
& T2RD & $\mathbf{857 \pm 12}$ & $\mathbf{847 \pm 11}$ & $\mathbf{725 \pm 37}$ \\
\bottomrule
\end{tabular}%
}
\label{tab:combined_data_augmentation}
\end{table}

\begin{table}
\caption{\textbf{Experimental results for different data augmentation sampling probability parameter \( p \).} We select ``Finger Spin'' task from DMControl-GB to conduct experiments under three environmental settings: ``Color Hard'', ``Video Easy'', and ``Video Hard''.}
\centering
\resizebox{0.65\textwidth}{!}{%
\begin{tabular}{l|c|c|c|c|c}
    \toprule
    Finger, & \multirow{2}{*}{$p = 0.1$} & \multirow{2}{*}{$p = 0.3$} & \multirow{2}{*}{$p = 0.5$} & \multirow{2}{*}{$p = 0.7$} & \multirow{2}{*}{$p = 0.9$} \\
    Spin & & & & &\\
    \midrule
    Color Hard & 968 $\pm$ 12 & 943 $\pm$ 11 & \textbf{984 $\pm$ 3} & 969 $\pm$ 19 & 975 $\pm$ 9 \\
    Video Easy & 719 $\pm$ 33 & 923 $\pm$ 18 & \textbf{966 $\pm$ 10} & 958 $\pm$ 17 & 961 $\pm$ 8 \\
    Video Hard & 459 $\pm$ 27 & 651 $\pm$ 49 & \textbf{833 $\pm$ 18} & 813 $\pm$ 18 & 703 $\pm$ 26 \\
    \bottomrule
\end{tabular}
}
\label{table:sensitive}
\end{table}

\begin{table}
\caption{\textbf{Hyperparameters} adopted in DMControl-GB and Robotic Manipulation tasks.}
\centering
\resizebox{0.90\textwidth}{!}{%
\begin{tabular}{l c c}
    \toprule
    \textbf{Hyperparameter} & \textbf{DMControl-GB} & \textbf{Robotic Manipulation}\\
    \midrule
    Frame rendering & $84 \times 84 \times 3$ & $84 \times 84 \times 3$ \\
    Frame stack & 3 & 1\\
    Action repeat & 8 (\textbf{Cartpole Swingup}), 2 (\textbf{Finger Spin}), 4 (otherwise) & 1 \\
    Random shift & $\pm 4$ pixels & $\pm 4$ pixels \\
    Discount factor $\gamma$ & 0.99 & 0.99 \\
    Episode length & 1000 & 50 \\
    Train steps & $0.5$M & $0.5$M \\
    Replay buffer size & $0.5$M & $0.5$M \\
    Optimizer ($\alpha$) & Adam(lr = $1\mathrm{e}{-4}$, $\beta_1$ = 0.5, $\beta_2$ = 0.999) & Adam(lr = $1\mathrm{e}{-4}$, $\beta_1$ = 0.5, $\beta_2$ = 0.999)\\
    Optimizer (RL) & Adam(lr = $1\mathrm{e}{-3}$, $\beta_1$ = 0.9, $\beta_2$ = 0.999) & Adam(lr = $1\mathrm{e}{-3}$, $\beta_1$ = 0.9, $\beta_2$ = 0.999)\\
    Optimizer (T2RD) & Adam(lr = $3\mathrm{e}{-4}$, $\beta_2$ = 0.999) & Adam(lr = $3\mathrm{e}{-4}$, $\beta_2$ = 0.999)\\
    Batch size (RL) & 128 & 128 \\
    Batch size (T2RD) & 128 & 256 \\
    Actor update freq. & 2 & 2 \\
    Critic target update freq. & 2 & 2 \\
    Aux update freq. & 2 & 2 \\
    EMA update coef. $\rho$ & 0.005 & 0.005 \\
    Aug sampling prob. $p$ & 0.5 & 0.5 \\
    Loss coefficient $\lambda_1$ & 1 & 1 \\
    Loss coefficient $\lambda_2$ & 100 & 100 \\
    Loss coefficient $\lambda_3$ & 0.1 & 0.1 \\
    \bottomrule
\end{tabular}
}
\label{table:hyperparameters}
\end{table}

\section{Reproducibility}
\textbf{Implementation Details.}
To ensure a fair comparison, we use the same hyperparameter settings as most methods, such as SODA and SVEA, in the DeepMind Control Suite. 
In Robotic Manipulation tasks, our hyperparameters follow those used in SGQN. 
The specific hyperparameter settings are detailed in Table~\ref{table:hyperparameters}.

\textbf{Network Architecture Details.}
We adopt a BYOL-inspired architecture for representation learning. The observation encoder \( f_{\theta} \) is an 11-layer CNN that maps stacked observations (9 $\times$ 84 $\times$ 84) to 14112-dimensional latent vectors, which are projected to 200 dimensions and then split into task-relevant and task-irrelevant features (100 each) by a dual-head MLP \( g_{\theta} \).
A predictor network \( h_{\theta} \), consisting of a linear layer with batch normalization and ReLU, encourages consistency between augmented and original task-relevant representations. 
To achieve decoupling, we use two auxiliary modules: (i) a cross-reconstruction network with two linear layers and one ReLU, which combines \(\mu_t'\) and \(\epsilon_t\) to reconstruct \(z_t\); and (ii) a cross-dynamic prediction network with three linear layers and two ReLUs, which predicts the next-step task-relevant features \(\mu_{t+1}\) from \(\mu_t'\) and action \(a_t\). 
For policy learning, we adopt SAC, where the actor and critic are standard three-layer MLPs with ReLU activations. The policy learner shares the encoder and projection with T2RD, ensuring that only task-relevant features are used for control. The SAC objective \(L_{\mathrm{RL}}\) and the self-supervised objective \(L_{\mathrm{T2RD}}\) are jointly optimized using Adam, with alternating updates every \(\omega\) steps.

\section{Conclusion}
We introduce the concept of decoupling observations into task-relevant and task-irrelevant features in Visual Reinforcement Learning (VRL), making the learned policy more robust to task-irrelevant visual changes.
Based on this idea, we propose Task-Relevant Representation Decoupling (T2RD), a self-supervised algorithm for VRL.
T2RD consists of three components: \textit{task-relevant representation consistency}, \textit{cross-reconstruction}, and \textit{cross-dynamic prediction}.
The first two components achieve decoupling of content and style features at the visual level, while the third part learns task-relevant representations through dynamic prediction.
In addition, we propose a simple yet effective combined data augmentation approach.
We conduct extensive experiments to verify the generalization ability and sample efficiency of T2RD across the DeepMind Control Suite and Robotic Manipulation tasks.

\section{Limitations and Future Work}

Despite the strong generalization performance of T2RD, we acknowledge several limitations that also point to future directions.


First, T2RD assumes visual observations can be decoupled into task-relevant ``content'' and task-irrelevant ``style'' streams. While this assumption holds in many benchmarks, it may fail when task-relevant cues are intertwined with style, such as navigation relying on specific terrain textures. In such cases, the model might treat crucial information as irrelevant. Future work could explore causal inference or adaptive mechanisms to dynamically identify task-relevant features.

Second, although T2RD is trained from scratch, its principle of decoupling remains highly relevant in the era of Visual Foundation Models (VFMs). Object-centric features extracted by VFMs still mix relevant states (e.g., pose) with irrelevant appearance (e.g., texture, lighting). Applying T2RD as a lightweight, self-supervised fine-tuning module to these features is a promising direction.

Finally, extending T2RD to real-world robotics is essential. 
The simulation-to-reality gap in lighting, textures, and object appearance often causes policy failures. 
Our decoupling approach is well-suited to mitigate this challenge, and validating T2RD on physical robots is a key future step.

\begin{acks}
This work was supported by the National Natural Science Foundation of China (No. 62576029) and the National Natural Science Foundation of China (No. 62536001).
\end{acks}











\end{document}